%% file: AAMAS_2026_sample.tex
\documentclass[sigconf,authorversion,nonacm]{aamas}

\usepackage{cleveref}
\usepackage{makecell}
\usepackage{multirow}
\usepackage{float}
\usepackage{pifont}
\usepackage{subcaption}
\usepackage{multirow}
\usepackage{balance} 
\newcommand{\acronym}{Harmanoid}

\setcopyright{ifaamas}
\acmConference[AAMAS '26]{Proc.\@ of the 25th International Conference
on Autonomous Agents and Multiagent Systems (AAMAS 2026)}{May 25 -- 29, 2026}
{Paphos, Cyprus}{C.~Amato, L.~Dennis, V.~Mascardi, J.~Thangarajah (eds.)}
\copyrightyear{2026}
\acmYear{2026}
\acmDOI{}
\acmPrice{}
\acmISBN{}

\title[AAMAS-2026 Formatting Instructions]{It Takes Two: Learning Interactive Whole-Body Control Between Humanoid Robots}

\author{Zuhong Liu\footnotemark[1]}
\affiliation{
  \institution{Shanghai Jiao Tong University}
  \city{Shanghai}
  \country{China}}
\email{albert_liu@sjtu.edu.cn}

\author{Junhao Ge\footnotemark[1]}
\affiliation{
  \institution{Shanghai Jiao Tong University}
  \city{Shanghai}
  \country{China}}
\email{cancaries@sjtu.edu.cn}

\author{Minhao Xiong}
\affiliation{
  \institution{Shanghai Jiao Tong University}
  \city{Shanghai}
  \country{China}}
\email{xmh123@sjtu.edu.cn}

\author{Jiahao Gu}
\affiliation{
  \institution{Shanghai Jiao Tong University}
  \city{Shanghai}
  \country{China}}
\email{gu1610755435@sjtu.edu.cn}

\author{Bowei Tang}
\affiliation{
  \institution{Shanghai Jiao Tong University}
  \city{Shanghai}
  \country{China}}

\author{Wei Jing}
\affiliation{
  \institution{Nerv.ai}
  \city{Shanghai}
  \country{China}}

\author{Siheng Chen}
\affiliation{
  \institution{Shanghai Jiao Tong University}
  \city{Shanghai}
  \country{China}}
\email{sihengc@sjtu.edu.cn}

\begin{abstract}
The true promise of humanoid robotics lies beyond single-agent autonomy: two or more humanoids must engage in physically grounded, socially meaningful whole-body interactions that echo the richness of human social interaction. However, single-humanoid methods suffer from the isolation issue, ignoring inter-agent dynamics and causing misaligned contacts, interpenetrations, and unrealistic motions. To address this, we present \acronym\ , a dual-humanoid motion imitation framework that transfers interacting human motions to two robots while preserving both kinematic fidelity and physical realism. \acronym\ comprises two key components: (i) contact-aware motion retargeting, which restores inter-body coordination by aligning SMPL contacts with robot vertices, and (ii) interaction-driven motion controller, which leverages interaction-specific rewards to enforce coordinated keypoints and physically plausible contacts. By explicitly modeling inter-agent contacts and interaction-aware dynamics, \acronym\ captures the coupled behaviors between humanoids that single-humanoid frameworks inherently overlook. Experiments demonstrate that \acronym\ significantly improves interactive motion imitation, surpassing existing single-humanoid frameworks that largely fail in such scenarios. Our code will be released at \url{https://github.com/ZuhongLIU/Harmanoid}.
\end{abstract}


\newcommand{\BibTeX}{\rm B\kern-.05em{\sc i\kern-.025em b}\kern-.08em\TeX}

\begin{document}

\pagestyle{fancy}
\fancyhead{}


\maketitle 

\renewcommand{\thefootnote}{\fnsymbol{footnote}}
\footnotetext[1]{These authors contribute equally to this work.}
\input{contents/introduction}

\input{contents/related_works}
\input{contents/method}
\input{contents/exp}

\input{contents/futurework}



\bibliographystyle{ACM-Reference-Format} 
\bibliography{sample}

\end{document}

%% file: contents/introduction.tex
\section{Introduction}
\label{sec:intro}
\begin{figure}[htbp]
    \centering
    \includegraphics[width=\linewidth]{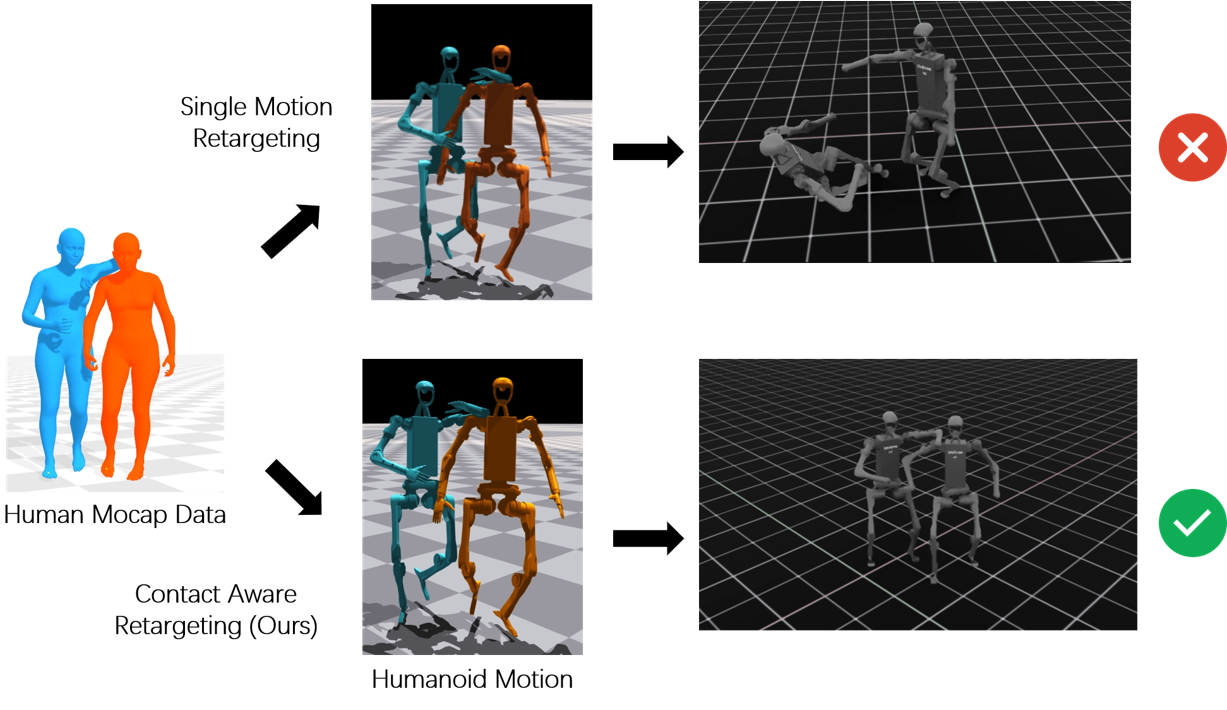} 
    \vspace{-8mm}
    \caption{In dual-humanoid scenarios, the isolation issue of single-humanoid motion imitation causes interpenetrations during motion retargeting, leading to instability and unrealistic behaviors in motion controller training.}
    \label{fig:problem}
    \vspace{-6mm}
\end{figure}

The full potential of humanoid robotics emerges when multiple robots can engage in physically grounded, socially meaningful interactions. These interactions encompass both coordination among robots and socially expressive behaviors between humans and robots. Potential applications range from elder care, rehabilitation support, and household assistance to industrial assembly and collaborative manufacturing. A key pathway toward this goal is humanoid motion imitation, which leverages abundant human-to-human interactions as learning data to enable robots to perform coordinated, physically realistic whole-body behaviors.

Current humanoid motion imitation research focuses on enabling humanoid robots to replicate diverse human behaviors, achieving increasingly expressive and physically realistic full-body motions across a wide range of tasks. Recent advances arise from two directions: data and algorithm. First, large-scale motion capture datasets provide rich human motion examples that serve as invaluable training data \cite{AMASS:ICCV:2019,xu2024inter,Liang_2024}. Second, improvements in humanoid motion imitation pipelines allow these motions to be retargeted to robot embodiments, refined through reinforcement learning in simulation, and ultimately deployed to real robots via sim-to-real techniques \cite{luo2023perpetualhumanoidcontrolrealtime,ze2025gmr,he2024omnih2o,yang2025omniretargetinteractionpreservingdatageneration,he2024omnih2o, ji2024exbody2, he2025hover,he2025asap, xie2025kungfubot,mittal2023orbit,makoviychuk2021isaac,tobin2017domain}.

Despite rapid advances in humanoid motion imitation, existing methods primarily focus on single humanoid scenarios and are inherently ill-suited for multi-humanoid interactions. A key challenge is the isolation issue, which arises when each humanoid is treated independently, neglecting the coupled dynamics and interdependencies that emerge during interactions. This issue causes several critical consequences: misaligned contacts and interpenetrations between agents, loss of coordinated behaviors in closely coupled motions such as hand-holding or shoulder contact, and physically implausible executions due to morphological mismatches, joint limitations, actuation constraints, and balance requirements. Together, these factors prevent current single-humanoid pipelines from faithfully reproducing interactive behaviors and highlight the need for approaches that explicitly model inter-agent dynamics.

\begin{figure*}[htbp]
    \vspace{-6mm}
    \centering
    \includegraphics[width=0.9\linewidth]{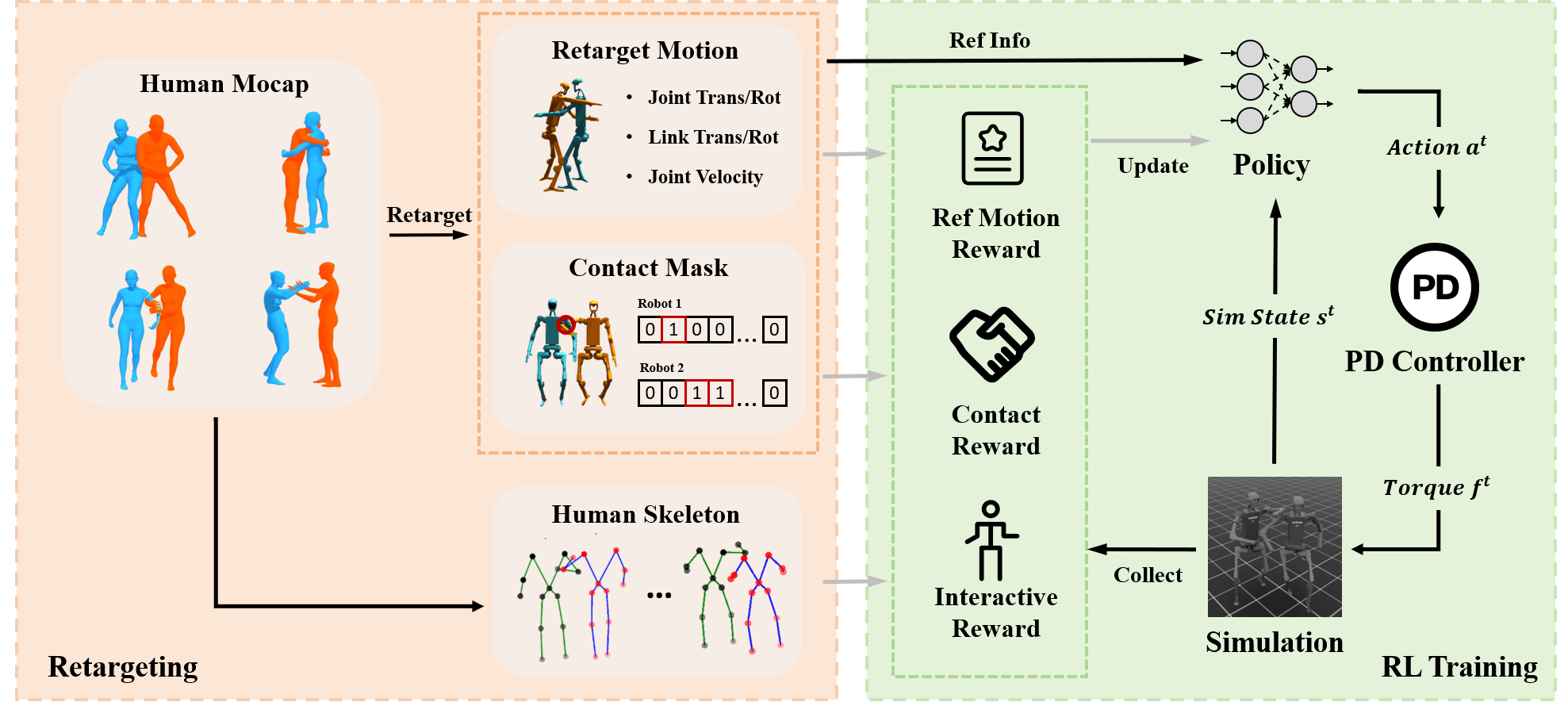} 
    \vspace{-2mm}
    \caption{Overview of the proposed \acronym\ framework. It contains two key components: (i) contact-aware motion retargeting, which optimize inter-body distance by aligning SMPL contacts with robot vertices, and (ii) interaction-driven motion controller, which leverages interaction-specific rewards to enforce coordinated keypoints and physically plausible contacts.}
    \label{fig:framework}
    \vspace{-4mm}
\end{figure*}

To address this, our key idea is to leverage contact information from mesh collision detection and interaction-aware guidance signals to enhance both kinematic fidelity and physical realism in humanoid motion imitation. Based on on this idea, we propose \acronym\ , the first dual-humanoid motion imitation framework that transfers the motions of two interacting humans to two humanoid robots while preserving both interaction coherence and physical plausibility; see Fig.~\ref{fig:framework}. \acronym\ comprises two key components: (i) a contact-aware motion retargeting module, which aligns human-to-robot contact correspondences to maintain realistic inter-body distances; and
(ii) an interaction-driven motion controller, which leverages interaction-specific reward design and curriculum learning to enforce coordinated whole-body behaviors. Compared with single-humanoid imitation methods, \acronym\ better handles the isolation issue in dual-humanoid interactions, achieving more realistic and physically feasible reference motions, smoother motion coordination, and enhanced physical feasibility.

In our experiments, we comprehensively evaluate the proposed dual-humanoid motion imitation framework, which integrates both motion retargeting and motion control. Compared to single humanoid imitation pipelines, our framework achieves significantly higher success rates and more accurate motion tracking, effectively handling scenarios where single-motion imitation largely fails, highlighting its ability in dealing with the isolation issue. For the motion retargeting stage, experiments demonstrate that our approach effectively preserves interactive characteristics after shape optimization and substantially mitigates undesired interpenetrations. It achieves around a 25\% improvement in success rate over state-of-the-art single motion retargeting methods. For the motion controller stage, ablation results confirm that incorporating interactive and contact-based rewards leads to more expressive and physically consistent behaviors. The use of curriculum learning further stabilizes training and refines the resulting motion quality.

In summary, our main contributions are as follows:
\vspace{-1mm}
\begin{itemize}
    \item We propose \acronym\ , the first dual-humanoid motion imitation framework that preserves both kinematic fidelity and physical realism in interactive motion imitation.

    \item \acronym\ introduces a two-stage pipeline, including a novel contact-aware motion retargeting module for inter-body coordination and a novel interaction-driven motion controller for physically plausible, coordinated behaviors.

    \item  \acronym\ achieves state-of-the-art performance over single motion imitation frameworks on Inter-X Dataset, enabling synchronized and stable interactive motions that  enhance humanoid robots’ collaborative capabilities.  
\end{itemize}

%% file: contents/related_works.tex
\section{Related Work}
\subsection{Motion Retargeting}
Motion retargeting addresses the fundamental challenge of transferring motion between morphologically distinct embodiments~\cite{yang2025omniretargetinteractionpreservingdatageneration, zhang2023simulationretargetingcomplexmulticharacter}, such as from humans to robots. Given a reference motion sequence from a source embodiment, motion retargeting produces a corresponding motion for the target embodiment that preserves the source's semantic content while respecting the target's physical constraints. Motion retargeting serves as a critical prerequisite for humanoid imitation learning by converting captured human motion data into reference trajectories that humanoids can track~\cite{he2024omnih2o, he2025asap, he2025hover}. Retargeting motions to humanoid robots introduce challenges lie in balancing similarity to the reference motion with physical plausibility for the target embodiment. PHC~\cite{luo2023perpetualhumanoidcontrolrealtime} jointly uses shape optimization and keypoint matching, treating motion retargeting as an optimization problem by minimizing weighted pose difference between selected keypoints from corresponding embodiment. Mink~\cite{Zakka_Mink_Python_inverse_2025} provides an inverse-kinematics based retargeting framework that uses quadratic programming to resolve kinematic constraints while matching reference keypoints. GMR~\cite{ze2025twist, ze2025gmr} mitigates artifacts such as foot sliding and ground penetration through non-uniform motion scaling and a two-stage optimization procedure. While these methods effectively transfer human motions into physically executable motions for humanoid robots, they primarily focus on single-agent motion expressiveness. When extended to dual-agent scenarios, they often fail to preserve critical interaction-specific properties, leading to uncoordinated behaviors such as unrealistic inter-robot distances and contact inconsistencies.

\subsection{Interactive Motion Synthesis}
Interactive motion synthesis in computer graphics has been extensively studied, with particular focus on coordination, multi-agent interactions, and the generation of physically plausible motions. The advent of deep learning has substantially advanced character animation \cite{barquero2024seamless,10.1145/3355089.3356505,chen2024pay,li2025unimotion}, enabling the creation of motions that are both more realistic and diverse. For human-object interactions, phase-function-based methods \cite{10.1145/3072959.3073663} allow characters to interact dynamically with objects, supporting tasks such as carrying a box \cite{xu2024regennethumanactionreactionsynthesis} or playing basketball \cite{10.1145/3386569.3392450}. These techniques have been further extended to handle a broader spectrum of static objects \cite{kulkarni2024nifty,rudin2022learningwalkminutesusing}, allowing characters to adapt their motion to complex and variable environments. In the domain of human-to-human interactions, most approaches rely on kinematic-based methods \cite{shafir2023human,ghosh2024remos} and datasets capturing multi-character interactions \cite{9156717,9093627,xu2024inter,Liang_2024}. Prior work has modeled behaviors such as pedestrian movement and collision avoidance in crowds \cite{10.1145/3424636.3426894}, while other studies explore coordinated multi-character interactions and object manipulations \cite{xu2024regennethumanactionreactionsynthesis}. \cite{gao2024coohoilearningcooperativehumanobject} address multi-humanoid object transportation through a two-phase learning paradigm: first acquiring individual skills, then transferring policies to enable cooperative tasks. However, these approaches remain limited to simulated characters and cannot extend to embodied humanoid robots due to strong physical constraints such as joint limits and actuation restrictions.

\subsection{Humanoid Motion Imitation}
In recent years, learning-based approaches have emerged as a powerful paradigm for achieving whole-body control in humanoid robots. In computer graphics, DeepMimic~\cite{2018-TOG-deepMimic} represents an early pioneering work demonstrating that incorporating reference motion data as imitation objectives enables characters to learn physically plausible control policies. With open-source kinematic retargeting pipelines~\cite{luo2023perpetualhumanoidcontrolrealtime, ze2025gmr, Zakka_Mink_Python_inverse_2025} that generate optimized trajectories for humanoid robots, there has been a surge of recent work focusing on motion imitation on humanoid robots, covering a board range of tasks such as whole body motion control~\cite{he2025asap, xie2025kungfubot}, teleoperation~\cite{he2024omnih2o, he2025hover} and loco-manipulation~\cite{yin2025visualmimicvisualhumanoidlocomanipulation, weng2025hdmilearninginteractivehumanoid,ben2025homiehumanoidlocomanipulationisomorphic}. GMT [17] improves robustness in tracking highly dynamic motions by prioritizing root velocity and local pose information over global positions. UniTracker\cite{yin2025unitracker} also supports dynamic movements, but its dependence on global position targets reduces stability when executing long motion sequences. BeyondMimic~\cite{liao2025beyondmimicmotiontrackingversatile} achieves state-of-the-art results on high-fidelity reference motions with minimal reward formulations. Despite this progress, current humanoid motion imitation approaches remain largely limited to single-agent settings. They do not adequately address multiple agent scenarios, where at least two humanoid robots jointly imitate interactive behaviors. To the best of our knowledge, our work is the first to address the problem of dual-humanoid motion imitation, enabling interactive motion controlling between two humanoid robots.

%% file: contents/method.tex
\section{Preliminaries}
\subsection{Problem Formulation}
\label{sec:problem_formulation}
 Given a pair of synchronized human motion capture sequences $(M^{h_1}_{1:T}, M^{h_2}_{1:T})$ representing the motions of two interacting humans $h_1, h_2$ over time, where each $M^{h_i}_{1:T}$ consists of keypoint translations $T^{h_i}_{1:T}$ and rotations $R^{h_i}_{1:T}$, the goal of dual-humanoid motion imitation is to generate a corresponding pair of physically plausible humanoid robot motions $(M^{r_1}_{1:T}, M^{r_2}_{1:T})$, with $r_i$ denoting the $i$-th humanoid robot. The generated motions should faithfully reproduce both the individual kinematics and the interactive behaviors observed in the original human motion pair. 

\subsection{Humanoid Motion Imitation}
Existing motion imitation frameworks~\cite{cheng2024expressivewholebodycontrolhumanoid,he2024omnih2o,he2025hover,he2025asap} focus mainly on single humanoid robot, typically following a two-stage design. In the first stage, motion retargeting converts human motion sequences $M^{h_1}_{1:T}$ into robot-reference motion $\widehat{M}^{r_1}_{1:T}$ that are consistent with the robot’s kinematic structure. In the second stage, a reinforcement learning (RL)-based motion controller takes the robot’s current state and the reference motion $\widehat{M}^{r_1}_{1:T}$ as input to produce corresponding joint actions. 

\noindent \textbf{Retargeting Human Motion for Humanoid}
\label{sec: single_retarget}
For a single humanoid robot $r$ with kinematics $\mathcal{K}_{r}$, a retargeting operator $\mathcal{R}_{\phi}$ maps the human motion sequence $M^{h}_{1:T}$ to a robot-reference sequence $\widehat{M}^{r}_{1:T}$. This mapping aligns keypoints to the robot's morphology under joint/actuation limits. We explicitly denote the reference sequence as
\begin{equation}
    \widehat{M}_{1:T}^{r} = (\widehat{\theta}_{1:T}^{r}, \widehat{p}_{1:T}^{r}, \widehat{\omega}_{1:T}^{r}, \widehat{v}_{1:T}^{r}),
\end{equation}
where $\widehat{\theta}_{1:T}$ denotes the 3D joint rotations and 
$\widehat{p}_{1:T}$ the corresponding 3D joint positions. From these, we derive the angular velocities $\widehat{\omega}_{1:T}$ and linear velocities $\widehat{v}_{1:T}$. While $\widehat{M}_{1:T}^{r}$ preserves the skeletal expressiveness of $M_{1:T}^{h}$, it inevitably loses important details from the human data that keypoints alone cannot capture.

\noindent \textbf{RL-based Motion Control.}
\label{sec: single_control}
The learning problem in single motion control is formulated as a goal conditioned reinforcement learning (RL) task for a Markov Decision Process (MDP) defined by the tuple 
\[
\mathcal{M} = \langle \mathcal{S}, \mathcal{A}, \mathcal{T}, \mathcal{R}, \gamma \rangle,
\]
where $\mathcal{S}$ is the state space, $\mathcal{A}$ is the action space, $\mathcal{T}: \mathcal{S} \times \mathcal{A} \rightarrow \mathcal{S}$ defines the transition dynamics, $\mathcal{R}: \mathcal{S} \times \mathcal{A} \rightarrow \mathbb{R}$ is the reward function, and $\gamma \in [0,1]$ is the discount factor. Each robot adopts a shared control policy $\pi_{\theta}$. 

At each time step $t$, the robot $r$ observes a state $o_{t}=(s^{\mathrm{prop}}_{t}, \widehat{m}_{t})$, where $s^{\mathrm{prop}}_{t}$ represents the robot's proprioceptive state, $\widehat{m}_t$  are per-frame targets derived from the reference motion $\widehat{M}^{r}$.

Given $o_{t}$, the policy $\pi_{\theta}$ outputs an action $a_{t} \in \mathcal{A}$, specifying target joint angles, which are then converted to torques through a PD controller. The reward function for robot at time $r_{t}$ computed as $r_{t} = \mathcal{R}\big(o_{t}, a_{t}\big)$:

The objective is to maximize the expected cumulative discounted reward for all robots:
\begin{equation}
    \max_{\pi_{\theta}} \; \mathbb{E}_{\pi_{\theta}} \left[\sum_{t=1}^{T} \gamma^{t-1} r_{t} \right],
\end{equation}
subject to the robot dynamics and physical constraints. However, directly applying this formulation to dual-humanoid motion imitation may fail because the controller lacks interactive guidance and information about the other robot in $o_t$, which are necessary for coordinated motions.

\subsection{Challenge: Isolation Issue}
When each humanoid is treated independently using a single-motion imitation pipeline, the isolation issue arises in two ways: (i) the retargeting from human to robot motion is done independently for each agent, causing all interactivity information in the human data to be lost in this stage. This amplifies morphological mismatches and can lead to severe issues such as undesired interpenetration between agents; (ii) without awareness of other agents’ information or joint collaboration during RL training, the overall motion execution will appear awkward and uncoordinated, even if each robot can accurately perform its own individual motion.

\section{\acronym\ Framework}
In this section, we present \acronym\ , a dual-humanoid motion imitation framework that transfers interacting human motions to two robots while preserving both kinematic fidelity and physical realism. \acronym\ integrates two stages: (i) contact-aware motion retargeting, which ensures that interactions between agents are preserved while minimizing penetration artifacts; and (ii) an interaction-driven motion controller, which further refines the retargeted motions using reinforcement learning, guided by both an interaction-aware reward and a contact-based reward. 

\subsection{Contact-Aware Motion Retargeting}
\label{sec:motion-retargeting}
\begin{figure}[htbp]
    \vspace{-3mm}
    \centering
    \includegraphics[width=\linewidth]{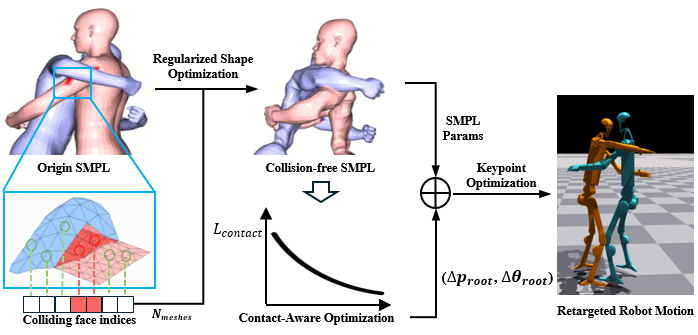}
    \vspace{-3mm}
    \caption{Overview of our motion retargeting pipeline that maps SMPL contacts to the robot mesh and optimizes root pose to ensure realistic, penetration-free motions.}
    \label{fig:retarget_draft}
    \vspace{-4mm}
\end{figure}

Discrepancies in body shape and proportions from single-agent retargeting methods, such as PHC \cite{luo2023perpetualhumanoidcontrolrealtime} or GMR \cite{ze2025gmr}, are amplified in interactive settings, leading to root pose deviations that cause collisions, penetrations, or unnatural distances between robots. To address this, our key idea is to use contact information as a powerful indicator of interactivity and ensure that this information should be well preserved in the retargeted humanoid motions. The retargeting is done through a three-step process: (i) mesh-based collision detection identifies contact vertices in the human motion; (ii) these vertices are mapped to the robot mesh obtained via regularized shape optimization; (iii) the relative root poses of the robots are optimized using the mapped contacts. These procedures restore interaction geometry and reduces penetration artifacts. Fig. \ref{fig:retarget_draft} demonstrates the overall procedure of the retargeting algorithm.

\noindent \textbf{Contact Mesh Detection. }
Formally, we represent a SMPL-format human motion sequence as
\begin{equation}
    M_{1:T}^{(i)} = (\beta^{(i)}, \theta^{(i)}_{1:T}, p^{(i)}_{1:T}),
\end{equation}
where $\beta^{(i)}$ denotes the body shape parameters, 
$\theta_{1:T}^{(i)} \in \mathbb{R}^{T \times 21 \times 3}$ are the body pose parameters, 
and $p^{(i)}_{1:T} \in \mathbb{R}^{T \times 3}$ are the root translations. 
The SMPL function $\mathcal{S}$ maps these parameters to a triangular mesh representation consisting of both vertices and faces:
\begin{equation}
{ V_{1:T}, F } = \mathcal{S}(\beta, \theta_{1:T}, p_{1:T})
, \quad
V_{1:T} \in \mathbb{R}^{T \times 6890 \times 3} ,
F \in \mathbb{N}^{13776 \times 3}
\end{equation}
where $V_{1:T}$ denotes mesh vertices coordinates, and $F$ represents the fixed face topology shared across frame, with each face defined by three vertex indices.

To capture the physical contacts between two humans, we perform per-frame mesh-face collision detection.
Let ${V_{t}^{(i)}, F^{(i)}}$ denote the per-frame SMPL mesh vertices and faces of person $i$ at time step $t$. We use the mesh collision detection algorithm in \texttt{Trimesh}~\cite{trimesh} to identify intersecting face pairs between the two human meshes:
\begin{equation}
\mathcal{C}_t = \text{CollisionPairs}(V_{t}^{(1)}, F^{(1)}, V_{t}^{(2)}, F^{(2)}),
\end{equation}
where
\begin{equation}
\mathcal{C}_t = \{(f_i, f_j) \mid f_i \in F^{(1)}, f_j \in F^{(2)}, \text{dist}(f_i, f_j) \le \epsilon \}
\end{equation}
is the set of colliding face pairs at frame $t$, indicating physical contact between corresponding body regions.

By performing this procedure for all frames $t = 1, \dots, T$, 
we obtain the full set of collision face pairs across the motion sequence:
\begin{equation}
    \mathcal{C}_{1:T} = \{\mathcal{C}_1, \mathcal{C}_2, \dots, \mathcal{C}_T\}.
\end{equation}

This information explicitly captures the contacts between humans, which can then be used to guide motion retargeting and enable contact-aware control of the robots.

\begin{figure}[htbp]
    \centering
    \includegraphics[width=0.95\linewidth]{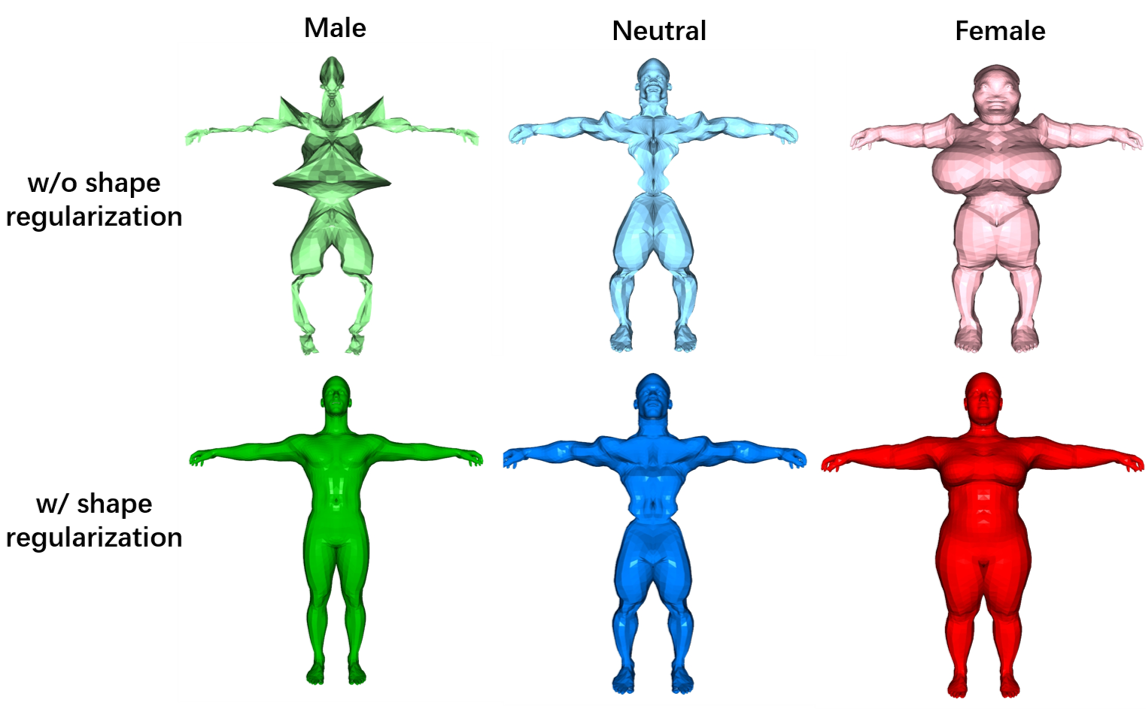}
    \vspace{-2mm}
    \caption{Comparison showing the effect of $\beta$ regularization during shape optimization, which keeps the human shape close to the robot structure.}
    \label{fig:smpl}
    \vspace{-5mm}
\end{figure}

\noindent \textbf{Regularized Shape Optimization }
To align the human body with the humanoid robot’s shape, a standard approach is to perform a similar fitting procedure as in \cite{luo2023perpetualhumanoidcontrolrealtime} to estimate the shape parameters $\beta'$ that best approximate the humanoid structure.  However, relying solely on keypoint matching- based shape optimization often leads to overfitting, resulting in unrealistic deformations of the human mesh (see Fig. \ref{fig:smpl}). From a mesh similarity standpoint, the optimized SMPL body may substantially deviate from the humanoid robot mesh, causing the corresponding vertices positions in collision pairs to drift notably from their intended locations on the robot surface. Consequently, the contact information becomes distorted after shape optimization, leading to a loss of interactivity during retargeting.

To mitigate this issue, we incorporate an $L_2$ regularization term on the shape parameters $\beta'$ into the retargeting loss:
\begin{equation}
    \mathcal{L}_{\text{retarget}} = \mathcal{L}_{\text{keypoint}} + \lambda \|\beta'\|_2^2,
\end{equation}
where $\mathcal{L}_{\text{keypoint}}$ denotes the keypoint position differences and $\lambda$ is the coefficient representing the strength of the regularization. This term encourages the optimized human shape to stay close to the humanoid robot’s structure at both the keypoint and mesh levels, preventing excessive deformation and maintaining mesh consistency for subsequent motion retargeting.

\noindent \textbf{Contact-Aware Root Pose Optimization. }
Ideally, the retargeted robot motions should replicate the same contact patterns observed in the human motion capture data, touching at corresponding body parts to preserve the interaction. To preserve inter-robot contact during motion retargeting, we introduce a learnable root pose offset for the first agent (without loss of generality) and optimize it to align corresponding contact vertices. Let $(v'_i, v'_j)$ denote a pair of colliding face centroids projected on the robot-shaped SMPL mesh with shape parameter $\beta'$ at frame $t$, where each centroid is computed as the mean of its three vertices:
\begin{equation}
\begin{aligned}
    \hat{v}_i &= \frac{1}{3}\sum_{k=1}^{3} \mathcal{S}(\beta', \theta^{(1)}_{t}, p_{t}^{(1)})[f_i^{(k)}], \\
    \hat{v}_j &= \frac{1}{3}\sum_{k=1}^{3} \mathcal{S}(\beta', \theta^{(2)}_{t}, p_{t}^{(2)})[f_j^{(k)}],
\end{aligned}
\end{equation}
$f_i^{(k)}$ and $f_j^{(k)}$ are contact vertex indices of faces $f_i$ and $f_j$ $\in \mathcal{C}_t$.

The root translation and rotation offsets $\Delta p_{root}, \Delta \theta_{root}$ are obtained by minimizing the sum of centroid position differences over the sequence:
\begin{equation}
    (\Delta p_{root}, \Delta \theta_{root}) = \arg\min \sum_{t=1}^T \sum_{(v_i, v_j) \in \mathcal{C}_t} \|\hat{v}_i - \hat{v}_j\|_2^2.
\end{equation}
This adaptively adjusts the relative pose between robots, reducing penetration or excessive separation at contact points. The updated root pose of the first SMPL is then:
\begin{equation}
    p_{root}^{(1)'} = p_{root}^{(1)} + \Delta p_{root} \in \mathbb{R}^3, \quad \theta_{root}^{(1)'} = \theta_{root}^{(1)} + \Delta \theta_{root} \in \mathbb{R}^3.
\end{equation}

We establish a joint mapping between human and robot and retarget motions by minimizing the L2 distance between corresponding joint positions, producing detailed kinematic reference trajectories $\widehat{\theta}_{1:T}^{(i)}, \widehat{p}_{1:T}^{(i)}, \widehat{\omega}_{1:T}^{(i)}, \widehat{v}_{1:T}^{(i)}$ for each robot. The per-frame contact mask $\widehat{c}_{t}^{(i)}$ has dimension equal to the number of robot links. Each element is set to 1 if the corresponding SMPL body region is in contact (according to $\mathcal{C}_t$), forming a one-hot vector marking active contact regions on the robot.

By automatically adjusting the optimal relative root poses that single motion retargeting pipeline fails to capture, our contact-aware motion retargeting generates more realistic and physically feasible robot reference motions.

\subsection{Interactive-Driven Motion Controller}
\label{sec:rl-motion-control}

After motion retargeting generates robot-reference sequences for each agent, we further refine the motions with reinforcement learning (RL) to achieve physically feasible and interaction-consistent behaviors. Directly optimizing a composite objective for tracking, coordination, and contact consistency is challenging due to nonconvex coupling and hard constraints. Instead, we train a shared policy that tracks the retargeted references while learning interaction-aware and contact-consistent behaviors via tailored rewards and an adaptive curriculum. The policy receives partner state summaries and reference contact masks, in addition to proprioception and reference targets, enabling actions synchronized with the partner and consistent with the contact schedule. A global interaction reward aligns pairwise upper-body keypoints with the reference, while contact reward encourage expected contacts and penalize unexpected ones using retargeted masks and measured forces. The online curriculum modulates the relative weighting of tracking, interaction, and contact terms, emphasizing reliable tracking early and progressively prioritizing interaction and contact. This design produces physically plausible motions that respect inter-agent geometry and contact timing without imposing brittle hard constraints.

\noindent \textbf{Interactive Observations.}
We extend the observation to supply the policy with interaction-relevant signals beyond proprioception and reference targets. For agent $i$ at time $t$, we augment the baseline inputs with: (i) a summary of the other agent’s current state $s^{\mathrm{other}}_{t}$ used to synchronize motion phase and relative upper-body geometry; (ii) per-link contact masks for both agents from the retargeted sequence, $\widehat{c}^{(i)}_{t}$ and $\widehat{c}^{(-i)}_{t}$, encoding intended inter-agent contacts; and (iii) the agent’s measured contact state $c^{(i)}_{t}$. Together with proprioception and reference targets $\widehat{m}^{(i)}_{t}$, these interaction signals drive actions to be coordinated with the partner and consistent with the contact schedule, mitigating single-agent isolation.

\noindent \textbf{Interaction-Aware Global Reward.}
To preserve the interaction patterns encoded in the human motion pair, we define a shared, global reward that encourages the pairwise spatial relationships between the two robots’ upper-body keypoints to match those of the reference. Let $p^{(i)}_{t,u} \in \mathbb{R}^{3}$ denote the 3D position of upper-body keypoint $u$ for agent $i$ at time $t$, and let $\widehat{p}^{(i)}_{t,u}$ be the corresponding reference. For two agents $(1,2)$ and selected keypoint sets, define
\begin{equation}
\begin{aligned}
\Delta^{\mathrm{sim}}_{t}(u,v) \;=\; p^{(1)}_{t,u} - p^{(2)}_{t,v}, \\
\Delta^{\mathrm{ref}}_{t}(u,v) \;=\; \widehat{p}^{(1)}_{t,u} - \widehat{p}^{(2)}_{t,v}.
\end{aligned}
\end{equation}
We compute reference-weighted pairwise importance
\begin{equation}
w_{t}(u,v) \;=\; \frac{\exp\!\big(-\|\Delta^{\mathrm{ref}}_{t}(u,v)\|_{2}/\sigma_{\mathrm{iw}}\big)}{\sum_{u',v'} \exp\!\big(-\|\Delta^{\mathrm{ref}}_{t}(u',v')\|_{2}/\sigma_{\mathrm{iw}}\big)},
\end{equation}
and a symmetric relative discrepancy
\begin{equation}
\label{eq: interactive discrepancy}
E_{t}(u,v) \;=\; \tfrac{1}{2}\,\frac{\|\Delta^{\mathrm{sim}}_{t}(u,v)-\Delta^{\mathrm{ref}}_{t}(u,v)\|_{2}}{\|\Delta^{\mathrm{sim}}_{t}(u,v)\|_{2}}
\;+\;
\tfrac{1}{2}\,\frac{\|\Delta^{\mathrm{sim}}_{t}(u,v)-\Delta^{\mathrm{ref}}_{t}(u,v)\|_{2}}{\|\Delta^{\mathrm{ref}}_{t}(u,v)\|_{2}}.
\end{equation}
The interaction reward is
\begin{equation}
r^{\mathrm{int}}_{t} \;=\; \exp\!\Big(-\sigma_{\mathrm{int}} \sum_{u,v} w_{t}(u,v)\, E_{t}(u,v)\Big),
\end{equation}
which is shared by both agents and rises when the simulated cross-agent offsets match the reference, smoothly penalizing relational mismatches and thereby coupling the policies to maintain inter-agent proximity and alignment.

\noindent \textbf{Contact Reward.}
We shape contact behavior using the reference contact masks $\widehat{c}^{(i)}_{t}$ and the measured contact forces on non-feet body links $b\in\mathcal{B}^{(i)}_{\mathrm{nf}}$ for agents $i\in\{1,2\}$. The contact-force magnitude is
$F^{(i)}_{t,b}=\big\|\mathbf{f}^{(i)}_{t,b}\big\|$.
We reward expected contacts and penalize unexpected contacts separately.
For expected-contact reward $r^{\mathrm{exp}}_{t,b,i}$, when the reference indicates contact ($\widehat{c}^{(i)}_{t,b}=1$), forces within $[f_{\min},f_{\max}]$ receive maximal reward; otherwise the reward decays sigmoidally with the distance to this interval:
\begin{equation}
r^{\mathrm{exp}}_{t,b,i}=
\widehat{c}^{(i)}_{t,b}\cdot
\begin{cases}
r_{\max}, & f_{\min}\le F^{(i)}_{t,b}\le f_{\max},\\[2pt]
\dfrac{r_{\max}}{1+\exp\!\big(\kappa\,(f_{\min}-F^{(i)}_{t,b})\big)}, & F^{(i)}_{t,b}<f_{\min},\\[8pt]
\dfrac{r_{\max}}{1+\exp\!\big(\kappa\,(F^{(i)}_{t,b}-f_{\max})\big)}, & F^{(i)}_{t,b}>f_{\max}.
\end{cases}
\end{equation}
For unexpected-contact penalty $p^{\mathrm{unexp}}_{t,b,i}$, when the reference indicates no contact ($\widehat{c}^{(i)}_{t,b}=0$), forces above a threshold are penalized with a sigmoidal increase in the excess over the threshold:
\begin{equation}
p^{\mathrm{unexp}}_{t,b,i}=\big(1-\widehat{c}^{(i)}_{t,b}\big)\cdot\frac{1}{1+\exp\!\big(-\kappa\,(F^{(i)}_{t,b}-\tau)\big)}.
\end{equation}
Aggregating over body links and agents yields the contact reward and penalty used in the curriculum:
\begin{equation}
r^{\mathrm{con}}_{t}=\sum_{i\in\{1,2\}}\sum_{b\in\mathcal{B}^{(i)}_{\mathrm{nf}}} r^{\mathrm{exp}}_{t,b,i},
\qquad
p^{\mathrm{con}}_{t}=\sum_{i\in\{1,2\}}\sum_{b\in\mathcal{B}^{(i)}_{\mathrm{nf}}} p^{\mathrm{unexp}}_{t,b,i}.
\end{equation}

\noindent \textbf{Curriculum between motion tracking and interaction.}
To mitigate early conflicts between goal-conditioned tracking and interaction objectives, we adopt an online curriculum that adjusts reward scales based on a normalized joint-velocity tracking proficiency. The total reward is
\begin{equation}
r_{t} \;=\; w_{\mathrm{trk}}(t)\, r^{\mathrm{goal}}_{t} \;+\; w_{\mathrm{int}}(t)\, r^{\mathrm{int}}_{t} \;+\; w_{\mathrm{con}}(t)\, r^{\mathrm{contact}}_{t} \;.
\end{equation}
Tracking proficiency is measured as
\begin{equation}
s_{t} \;=\; \frac{\frac{1}{|\mathcal{A}|}\sum_{i\in\mathcal{A}} r^{\mathrm{vel}}_{t,i}}{\gamma^{\mathrm{vel}}_{t}},
\end{equation}
where $r^{\mathrm{vel}}_{t,i}$ is the per-agent joint-velocity tracking reward and $\gamma^{\mathrm{vel}}_{t}$ is its current scale, with $\mathcal{A}$ indexing the controlled agents. A piecewise gain drives scale updates:
\begin{equation}
\alpha_{t} \;=\;
\begin{cases}
1, & s_{t}<1,\\[4pt]
\dfrac{1}{\sqrt{s_{t}}}, & s_{t}\ge 1
\end{cases}
\end{equation}
Let $\mathcal{T}$ denote tracking-related terms and $\mathcal{I}$ denote interaction/contact terms. We update  scales online as
\begin{equation}
\gamma_{k}(t^{+}) \;=\; \alpha_{t}\,\gamma_{k}(t)\quad \forall\, k\in\mathcal{T},
\qquad
\gamma_{m}(t^{+}) \;=\; \alpha_{t}^{-1}\,\gamma_{m}(t)\quad \forall\, m\in\mathcal{I},
\end{equation}
If $s_{t}<1$, scales remain fixed; once $s_{t}\ge 1$, tracking terms are downweighted by $1/\sqrt{s_{t}}$ while interaction/contact terms are upweighted by $\sqrt{s_{t}}$, emphasizing interaction after sufficient tracking proficiency.

Compared with per-agent trackers that treat the partner as disturbance, our interaction-driven controller explicitly couples the two robots, producing synchronized, physically consistent whole-body behaviors.

%% file: contents/exp.tex
\begin{figure*}[h]
    \centering
    \includegraphics[width=0.9\linewidth]{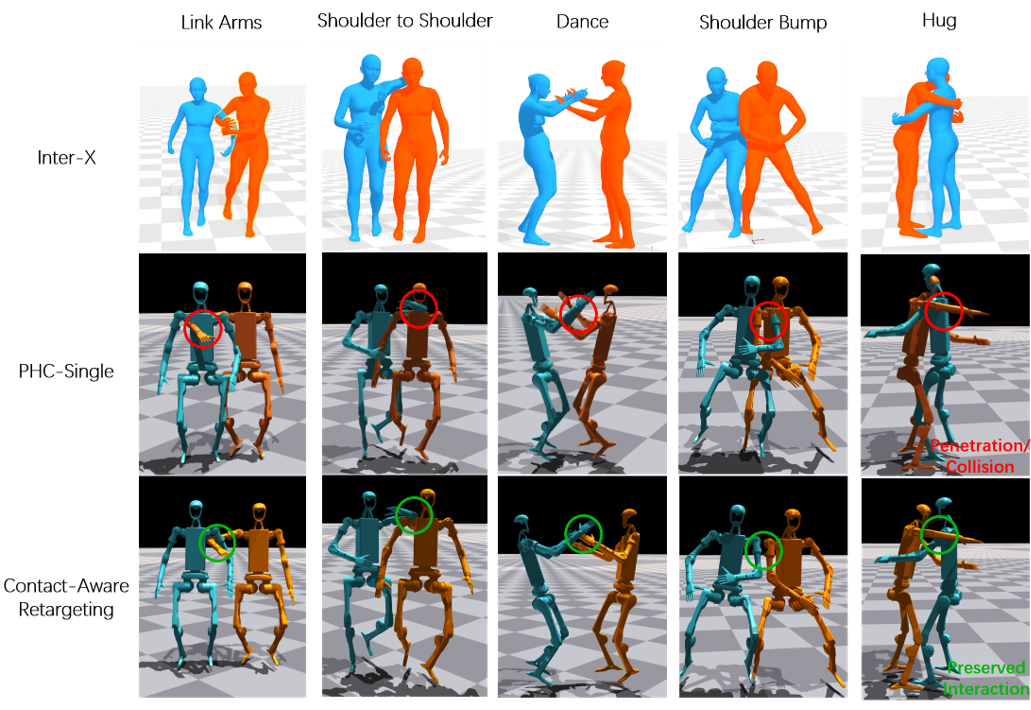} 
    \vspace{-5mm}
    \caption{Qualitative comparison between Inter-X mocap data, single-PHC retargeting, and our contact-aware retargeting algorithm. The \textcolor{red}{red} and \textcolor{green}{green} circles highlight regions of penetration/collision and successful avoidance, respectively. Our method better preserves interaction characteristics of the reference motion while reducing collisions.}
    \label{fig:retarget}
    \vspace{-2mm}
\end{figure*}

\section{Experiments}
In this section, we evaluate \acronym\ for dual-humanoid motion imitation to demonstrate its overall effectiveness. In~\cref{sec: exp_settings}, we describe our simulation setup. In ~\cref{sec:exp_motion_imitation}, we compare \acronym\ with single motion imitation baselines to highlight its advantages over existing approaches. ~\cref{sec: exp_motion_retargeting} details the motion retargeting evaluation, while ~\cref{sec: exp_motion_control} introduces the motion controller evaluation, emphasizing coordination and interactive behavior between robots.

\subsection{Experimental Settings}
\label{sec: exp_settings}
\noindent \textbf{Dataset.} We use the Inter-X dataset \cite{xu2024inter} as the source of our human-to-human motion capture data, which contains 11,388 interaction sequences with over 8.1M frames across 40 daily interaction categories. Since most interaction categories inherently belong to distant or contact-less interactions (e.g., waving, chatting) that can be reproduced by independently controlling two robots, we select around 50 representative sequences with strong interaction and physical contact, including various linked arms, shoulder-to-shoulder, hugging, dancing actions for our experiments.

\noindent \textbf{Implementation Details}
We adopt the Proximal Policy Optimization (PPO) algorithm \cite{schulman2017proximal}, following the implementation provided in \cite{rudin2022learningwalkminutesusing}, to train our control policies. All experiments are conducted in the IsaacSim simulator with a simulation time step of $\Delta t = 0.005$ seconds. The humanoid structure used in our study is the Unitree H1 robot, which features 19 degrees of freedom (DOFs).  A complete list of training hyperparameters is provided in the Appendix.

\noindent \textbf{Metrics.} To evaluate both the policy’s ability to maintain balance and its tracking performance, we adopt the following metrics: \textit{Success Rate}: a rollout is considered successful if the robot completes the episode without its anchor body height or orientation deviating from the reference beyond a predefined threshold; $E_{g\text{-}MPJPE}$ (mm), the global body position tracking error; $E_{MPJPE}$ (mm), the root-relative mean per-joint position error; $E_{acc}$ (mm/frame$^2$), the acceleration error; and $E_{vel}$ (mm/frame), the root velocity error.

\subsection{Dual Motion Imitation Evaluation}
\label{sec:exp_motion_imitation}
\noindent \textbf{Settings}. We compare our method with Exbody \cite{cheng2024expressivewholebodycontrolhumanoid} and the teacher motion imitator from HOVER \cite{he2025hover}, both of which have been shown to achieve stable and expressive single humanoid motion imitation. For motion retargeting, \cite{cheng2024expressivewholebodycontrolhumanoid} uses a straightforward approach that directly copies joint rotations from the source human motions to the target humanoid, while \cite{he2025hover} employs PHC\cite{luo2023perpetualhumanoidcontrolrealtime}, the most widely used retargeting in humanoid motion imitation\cite{he2024omnih2o,he2025hover,he2025asap,xie2025kungfubot}. Notably, both methods rely solely on proprioceptive information from a single robot during the motion control stage. For both baselines, we independently retarget each human motion using their original methods, train the motion controller on the combined motion datasets from both robots, and evaluate it by rolling out paired motions from the combined dual-robot sequences.

\noindent \textbf{Results}. Table \ref{tab:motion imitation metrics} illustrates the quantitative results. Notably, single motion imitation baselines completely fail to reproduce interactive behaviors, highlighting their limitations in dual-humanoid scenarios. Notably, we observe that single motion imitators can hardly succeed in performing a complete motion sequence in the simulation mainly due to collision between robots during the rollout. This phenomenon further shows the existence of the isolation issue, especially the amplified morphological misalignment problem that generates physically infeasible motions during retargeting.  In comparison, our method significantly outperforms these baselines, demonstrating both stable balance and synchronized motion control in interactive settings.

\begin{table}[t]
\centering
\setlength{\tabcolsep}{1mm}{
\begin{tabular}{l|ccccc}
\toprule
\textbf{Method} & \textbf{Succ$\uparrow$} & \textbf{$E_{g\text{-}MPJPE}\downarrow$} & \textbf{$E_{MPJPE}\downarrow$} & \textbf{$E_{Acc}\downarrow$} & \textbf{$E_{Vel}\downarrow$} \\
\midrule
Exbody\cite{cheng2024expressivewholebodycontrolhumanoid} & 0.00 & 193.6 & 91.7 & 5.41 & 9.43 \\
HOVER\cite{he2025hover} & 0.20 & 187.4 & 94.6 & 4.74 & 9.55 \\
Harmonoid (Ours) & \textbf{0.92} & \textbf{156.8} & \textbf{88.7} & \textbf{3.08} & \textbf{7.55} \\
\bottomrule
\end{tabular}
}
\caption{Comparison between \acronym\ and single motion imitation baselines across multiple metrics. Our method achieves the highest success rate and lowest errors, outperforming all single motion imitation baselines.}
\label{tab:motion imitation metrics}
\vspace{-8mm}
\end{table}

\subsection{Motion Retargeting Evaluation}
\label{sec: exp_motion_retargeting}
\noindent \textbf{Settings.}
We evaluate different interactive motion retargeting methods using retargeted motions executed by a motion controller trained on AMASS~\cite{AMASS:ICCV:2019}, following the single PHC retargeting and training procedure of~\cite{he2025hover}. Specifically, for each robot, we deploy the trained single motion controller to track the reference motion obtained from motion retargeting. Intuitively, interpenetrations between robots indicate conflicts in the reference motions; in such cases, a single motion controller is likely to fail in accurately tracking the interactive motions, potentially resulting in loss of balance due to physical collisions. This allows us to assess whether the retargeted motions are physically executable and reasonable, thereby validating the quality of the retargeting results.

For comparison, we consider the following three methods:
\begin{itemize}
\item \textbf{Single PHC.} Same setting with ~\cref{sec:exp_motion_imitation}.
\item \textbf{Single PHC + IC.} We extend PHC by adding an interaction constraint similar to Eq.~\ref{eq: interactive discrepancy} into the objective function for keypoint optimization, aiming to encourage correction for interactive behaviors during the retargeting stage.
\item \textbf{Harmanoid-Retarget.} Our proposed contact-aware motion retargeting method, which introduces contact awareness during retargeting by applying a regularized shape optimization with a hyperparameter $\beta = 0.0016$.
\end{itemize}

\begin{figure}[htbp]
    \vspace{-4mm}
    \centering
    \includegraphics[width=0.9\linewidth]{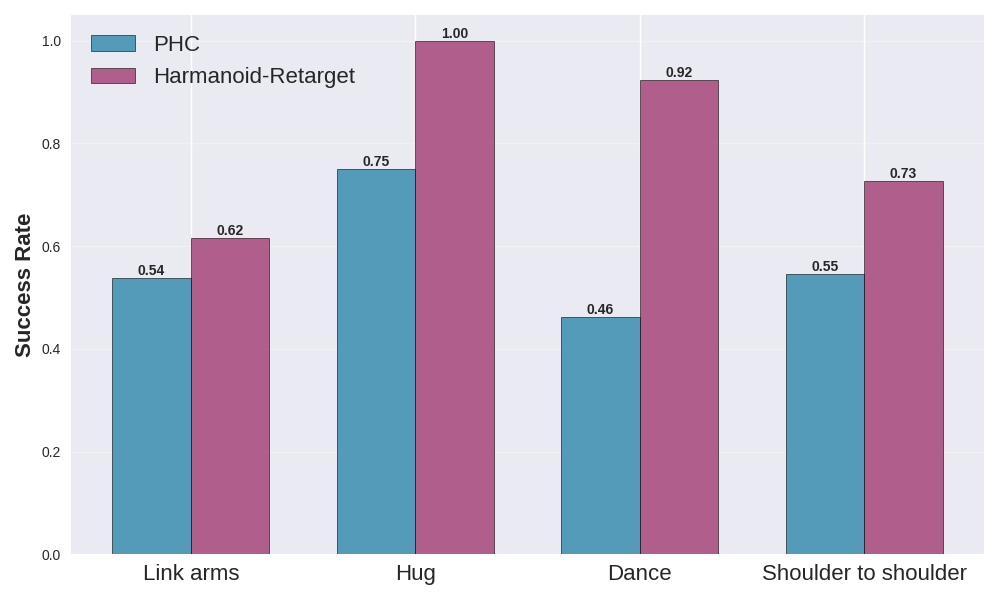} 
    \caption{Quantitative comparison of motion retargeting performance between Single PHC and \acronym\ - Retarget. Ours consistently achieves higher success rate across all types of interactive motions.}
    \label{fig:motion type}
    \vspace{-3mm}
\end{figure}

\noindent \textbf{Results.}
Table \ref{tab:retarget_metrics} shows the quantitative comparison between single motion imitation methods and \acronym . The Single PHC baseline shows lower tracking performance and success rate, mainly due to unrealistic reference motions that cause interpenetration. Adding a constraint to enforce interaction alone does not improve retargeting quality; in fact, it worsens tracking performance, mainly because it creates a trade-off between interaction fidelity and tracking accuracy, often compromising the upper-body pose expressiveness of a single robot. Consequently, the resulting motions deviate significantly from the motion distribution used to train the controller. In contrast, our method consistently improves both success rate and tracking accuracy, producing realistic and physically plausible motion transfers. Figure \ref{fig:motion type} shows a more detailed comparison across different motion categories. Our method consistently achieves higher success rates across all motion types compared with the single baseline, demonstrating its robustness and generality. Figure \ref{fig:retarget} qualitatively compares Single PHC and our retargeting method, showing that our approach better preserves interactivity and reduces interpenetration in different motion types.

\begin{table}[t]
\centering
\setlength{\tabcolsep}{1.0mm}{
\begin{tabular}{l|ccccc}
\toprule
\textbf{Method} & \textbf{Succ$\uparrow$} & \textbf{$E_{g\text{-}MPJPE}\downarrow$} & \textbf{$E_{MPJPE}\downarrow$} & \textbf{$E_{Acc}\downarrow$} & \textbf{$E_{Vel}\downarrow$} \\
\midrule
Single PHC & 0.62 & 170.8 & 78.4 & 4.68 & 10.6 \\
Single PHC + IC & 0.68 & 215.8 & 86.4 & 5.02 & 11.2 \\
Harmanoid-Retarget & \textbf{0.74} & \textbf{157.6} & \textbf{73.8} & \textbf{4.65} & \textbf{10.4} \\
\bottomrule
\end{tabular}
}
\caption{Comparison of retargeting methods across different evaluation metrics. Our method achieves better balancing and tracking performance compared with single motion retargeting pipeline.}
\label{tab:retarget_metrics}
\vspace{-8mm}
\end{table}

\subsection{Motion Control Evaluation}
\label{sec: exp_motion_control}
\noindent \textbf{Settings.} We evaluate our interaction-driven motion controller by training it with reference motions obtained from \acronym\ -Retarget. To assess the contributions of our proposed components, we perform an ablation study that incorporates additional rewards and curriculum learning to examine how they refine humanoid interactive behaviors. For evaluation, we use the success rate to measure overall task performance and the Contact F1 Score, defined as the F1 score between the predicted and ground-truth contact masks over the entire sequence, serving as a proxy for how well the robot follows the contact patterns of the original human data and reflecting the quality of interactivity between robots.

\begin{figure}[htbp]
    \centering
    \includegraphics[width=0.9\linewidth]{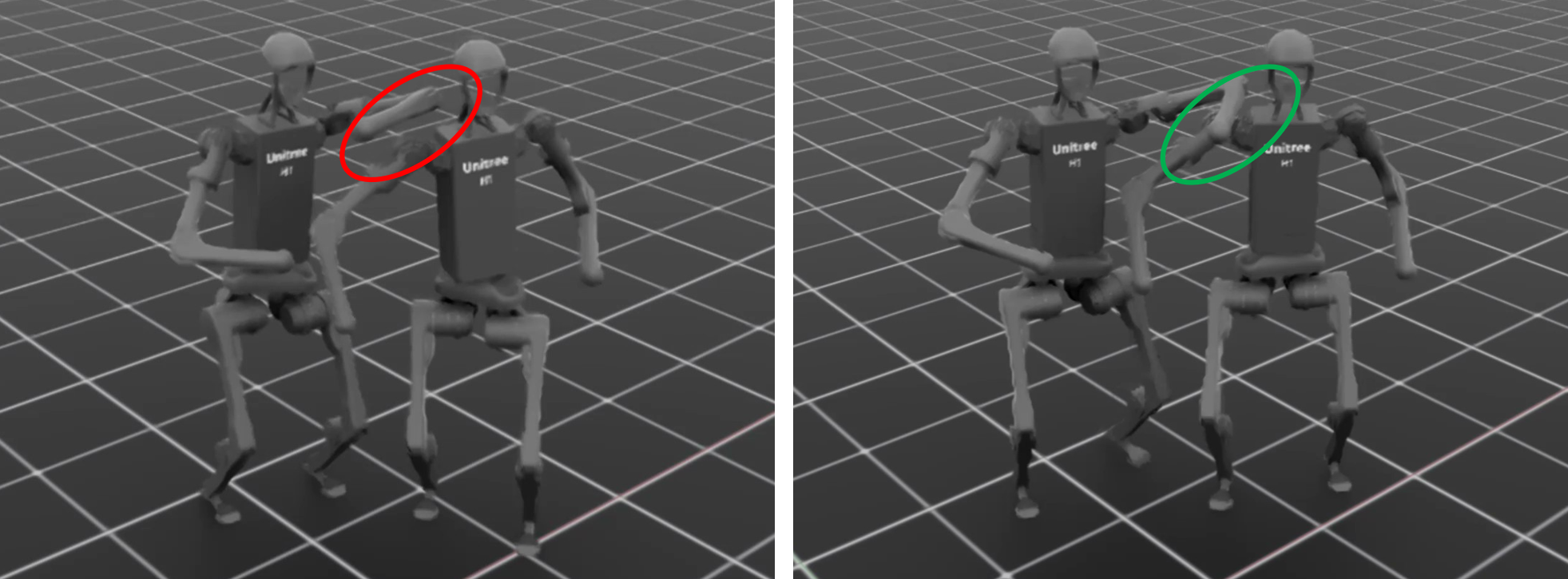} 
    \caption{Qualitative comparison between motion controllers with and without the additional supervised reward. The \textcolor{red}{red} and \textcolor{green}{green} circles indicate regions where incorporating the additional reward enables the policy to better present human-like interactive behaviors compared to the baseline.}
    \label{fig:motion control}
    \vspace{-4mm}
\end{figure}

\noindent \textbf{Results.} Tab. \ref{tab:ablation_multi_h1_motion} presents the ablation study from vanilla single policy training to the full interactive-driven motion controller on Inter-X. Adding our proposed components leads to consistent improvements in both task success rate and contact consistency. Incorporating the interaction reward enhances performance by encouraging more coordinated human-like behaviors between agents, while the contact reward further refines the quality of physical interactions. When all components, including the curriculum learning strategy, are combined, the controller achieves the most stable and realistic motions, demonstrating the complementary benefits of interaction modeling, contact awareness, and progressive training. Fig. \ref{fig:motion control} illustrates a qualitative comparison between motion controllers trained with and without the additional rewards. With the integration of interactive and contact rewards, the two humanoid robots better preserve the physical contact features and exhibit more coordinated behaviors in shoulder-to-shoulder actions.

\begin{table}[t]
\small
\centering
\setlength{\tabcolsep}{0.6mm}{
\begin{tabular}{l|ccc|cc}
\toprule
\textbf{Method} & \textbf{Inter. Rew.} & \textbf{Cont. Rew.} & \textbf{Curr.} & \textbf{Success Rate $\uparrow$} & \textbf{Contact F1 $\uparrow$} \\
\midrule
1 &  &  & & 0.84 & 0.10 \\
2 & $\checkmark$ &  &  & 0.88 & 0.16 \\
3 &  & $\checkmark$ &  & 0.90 & 0.15 \\
4 & $\checkmark$ & $\checkmark$ & $\checkmark$ & \textbf{0.92} & \textbf{0.18} \\
\bottomrule
\end{tabular}
}
\caption{Ablation study on the effectiveness of different components in \acronym\ . Inter.Rew: Interactive Reward. Cont.Rew: Contact Reward. Curr.: Curriculum. }
\label{tab:ablation_multi_h1_motion}
\vspace{-7mm}
\end{table}

%% file: contents/futurework.tex
\section{Discussions and Future Work}
In this work, we presented \acronym, a dual-humanoid motion imitation framework designed to enable physically grounded and socially meaningful interactions between humanoid robots. Unlike traditional single-humanoid imitation methods that suffers from isolation issue, \acronym\ explicitly models contact and coordination between interacting agents. By integrating contact-aware motion retargeting with interaction-driven motion control, our approach bridges the gap between kinematic realism and physical feasibility in multi-humanoid scenarios. The retargeting module preserves inter-body coordination through mesh-level contact alignment, while the controller enforces physically consistent and expressive behaviors via interaction-specific rewards and curriculum learning. Experimental results demonstrate that \acronym\ achieves substantially improved success rates, reduced interpenetrations, and more synchronized interactive behaviors compared with single-humanoid baselines. For future work, we aim to extend our framework to more challenging interactive scenarios, such as carrying a partner, which involve complex dynamic coordination. Additionally, deploying \acronym\ on real-world robots will require incorporating perception and communication mechanisms to share state information across agents, further bridging the gap between simulation and physical deployment.

%% file: AAMAS_2026_sample.bbl

\begin{thebibliography}{41}


\ifx \showCODEN    \undefined \def \showCODEN     #1{\unskip}     \fi
\ifx \showDOI      \undefined \def \showDOI       #1{#1}\fi
\ifx \showISBNx    \undefined \def \showISBNx     #1{\unskip}     \fi
\ifx \showISBNxiii \undefined \def \showISBNxiii  #1{\unskip}     \fi
\ifx \showISSN     \undefined \def \showISSN      #1{\unskip}     \fi
\ifx \showLCCN     \undefined \def \showLCCN      #1{\unskip}     \fi
\ifx \shownote     \undefined \def \shownote      #1{#1}          \fi
\ifx \showarticletitle \undefined \def \showarticletitle #1{#1}   \fi
\ifx \showURL      \undefined \def \showURL       {\relax}        \fi
\providecommand\bibfield[2]{#2}
\providecommand\bibinfo[2]{#2}
\providecommand\natexlab[1]{#1}
\providecommand\showeprint[2][]{arXiv:#2}

\bibitem[\protect\citeauthoryear{Barquero, Escalera, and Palmero}{Barquero et~al\mbox{.}}{2024}]%
        {barquero2024seamless}
\bibfield{author}{\bibinfo{person}{German Barquero}, \bibinfo{person}{Sergio Escalera}, {and} \bibinfo{person}{Cristina Palmero}.} \bibinfo{year}{2024}\natexlab{}.
\newblock \showarticletitle{Seamless human motion composition with blended positional encodings}. In \bibinfo{booktitle}{\emph{Proceedings of the IEEE/CVF Conference on Computer Vision and Pattern Recognition}}. \bibinfo{pages}{457--469}.
\newblock


\bibitem[\protect\citeauthoryear{Ben, Jia, Zeng, Dong, Lin, and Pang}{Ben et~al\mbox{.}}{2025}]%
        {ben2025homiehumanoidlocomanipulationisomorphic}
\bibfield{author}{\bibinfo{person}{Qingwei Ben}, \bibinfo{person}{Feiyu Jia}, \bibinfo{person}{Jia Zeng}, \bibinfo{person}{Junting Dong}, \bibinfo{person}{Dahua Lin}, {and} \bibinfo{person}{Jiangmiao Pang}.} \bibinfo{year}{2025}\natexlab{}.
\newblock \bibinfo{title}{HOMIE: Humanoid Loco-Manipulation with Isomorphic Exoskeleton Cockpit}.
\newblock
\newblock
\showeprint[arxiv]{2502.13013}~[cs.RO]
\urldef\tempurl%
\url{https://arxiv.org/abs/2502.13013}
\showURL{%
\tempurl}


\bibitem[\protect\citeauthoryear{Chen, Lu, Dai, Dou, Ju, Wang, Komura, and Zhang}{Chen et~al\mbox{.}}{2024}]%
        {chen2024pay}
\bibfield{author}{\bibinfo{person}{Ling-Hao Chen}, \bibinfo{person}{Shunlin Lu}, \bibinfo{person}{Wenxun Dai}, \bibinfo{person}{Zhiyang Dou}, \bibinfo{person}{Xuan Ju}, \bibinfo{person}{Jingbo Wang}, \bibinfo{person}{Taku Komura}, {and} \bibinfo{person}{Lei Zhang}.} \bibinfo{year}{2024}\natexlab{}.
\newblock \showarticletitle{Pay attention and move better: Harnessing attention for interactive motion generation and training-free editing}.
\newblock \bibinfo{journal}{\emph{arXiv preprint arXiv:2410.18977}} (\bibinfo{year}{2024}).
\newblock


\bibitem[\protect\citeauthoryear{Cheng, Ji, Chen, Yang, Yang, and Wang}{Cheng et~al\mbox{.}}{2024}]%
        {cheng2024expressivewholebodycontrolhumanoid}
\bibfield{author}{\bibinfo{person}{Xuxin Cheng}, \bibinfo{person}{Yandong Ji}, \bibinfo{person}{Junming Chen}, \bibinfo{person}{Ruihan Yang}, \bibinfo{person}{Ge Yang}, {and} \bibinfo{person}{Xiaolong Wang}.} \bibinfo{year}{2024}\natexlab{}.
\newblock \bibinfo{title}{Expressive Whole-Body Control for Humanoid Robots}.
\newblock
\newblock
\showeprint[arxiv]{2402.16796}~[cs.RO]
\urldef\tempurl%
\url{https://arxiv.org/abs/2402.16796}
\showURL{%
\tempurl}


\bibitem[\protect\citeauthoryear{{Dawson-Haggerty et al.}}{{Dawson-Haggerty et al.}}{[n.d.]}]%
        {trimesh}
\bibfield{author}{\bibinfo{person}{{Dawson-Haggerty et al.}}} \bibinfo{year}{[n.d.]}\natexlab{}.
\newblock \bibinfo{booktitle}{\emph{trimesh}}.
\newblock
\urldef\tempurl%
\url{https://trimesh.org/}
\showURL{%
\tempurl}


\bibitem[\protect\citeauthoryear{Fieraru, Zanfir, Oneata, Popa, Olaru, and Sminchisescu}{Fieraru et~al\mbox{.}}{2020}]%
        {9156717}
\bibfield{author}{\bibinfo{person}{Mihai Fieraru}, \bibinfo{person}{Mihai Zanfir}, \bibinfo{person}{Elisabeta Oneata}, \bibinfo{person}{Alin-Ionut Popa}, \bibinfo{person}{Vlad Olaru}, {and} \bibinfo{person}{Cristian Sminchisescu}.} \bibinfo{year}{2020}\natexlab{}.
\newblock \showarticletitle{Three-Dimensional Reconstruction of Human Interactions}. In \bibinfo{booktitle}{\emph{2020 IEEE/CVF Conference on Computer Vision and Pattern Recognition (CVPR)}}. \bibinfo{pages}{7212--7221}.
\newblock
\urldef\tempurl%
\url{https://doi.org/10.1109/CVPR42600.2020.00724}
\showDOI{\tempurl}


\bibitem[\protect\citeauthoryear{Gao, Wang, Xiao, Wang, Wang, Cao, Hu, Liu, Dai, and Pang}{Gao et~al\mbox{.}}{2024}]%
        {gao2024coohoilearningcooperativehumanobject}
\bibfield{author}{\bibinfo{person}{Jiawei Gao}, \bibinfo{person}{Ziqin Wang}, \bibinfo{person}{Zeqi Xiao}, \bibinfo{person}{Jingbo Wang}, \bibinfo{person}{Tai Wang}, \bibinfo{person}{Jinkun Cao}, \bibinfo{person}{Xiaolin Hu}, \bibinfo{person}{Si Liu}, \bibinfo{person}{Jifeng Dai}, {and} \bibinfo{person}{Jiangmiao Pang}.} \bibinfo{year}{2024}\natexlab{}.
\newblock \bibinfo{title}{CooHOI: Learning Cooperative Human-Object Interaction with Manipulated Object Dynamics}.
\newblock
\newblock
\showeprint[arxiv]{2406.14558}~[cs.RO]
\urldef\tempurl%
\url{https://arxiv.org/abs/2406.14558}
\showURL{%
\tempurl}


\bibitem[\protect\citeauthoryear{Ghosh, Dabral, Golyanik, Theobalt, and Slusallek}{Ghosh et~al\mbox{.}}{2024}]%
        {ghosh2024remos}
\bibfield{author}{\bibinfo{person}{Anindita Ghosh}, \bibinfo{person}{Rishabh Dabral}, \bibinfo{person}{Vladislav Golyanik}, \bibinfo{person}{Christian Theobalt}, {and} \bibinfo{person}{Philipp Slusallek}.} \bibinfo{year}{2024}\natexlab{}.
\newblock \showarticletitle{Remos: 3d motion-conditioned reaction synthesis for two-person interactions}. In \bibinfo{booktitle}{\emph{European Conference on Computer Vision}}. Springer, \bibinfo{pages}{418--437}.
\newblock


\bibitem[\protect\citeauthoryear{Haworth, Berseth, Moon, Faloutsos, and Kapadia}{Haworth et~al\mbox{.}}{2020}]%
        {10.1145/3424636.3426894}
\bibfield{author}{\bibinfo{person}{Brandon Haworth}, \bibinfo{person}{Glen Berseth}, \bibinfo{person}{Seonghyeon Moon}, \bibinfo{person}{Petros Faloutsos}, {and} \bibinfo{person}{Mubbasir Kapadia}.} \bibinfo{year}{2020}\natexlab{}.
\newblock \showarticletitle{Deep Integration of Physical Humanoid Control and Crowd Navigation}. In \bibinfo{booktitle}{\emph{Proceedings of the 13th ACM SIGGRAPH Conference on Motion, Interaction and Games}} (Virtual Event, SC, USA) \emph{(\bibinfo{series}{MIG '20})}. \bibinfo{publisher}{Association for Computing Machinery}, \bibinfo{address}{New York, NY, USA}, Article \bibinfo{articleno}{15}, \bibinfo{numpages}{10}~pages.
\newblock
\showISBNx{9781450381710}
\urldef\tempurl%
\url{https://doi.org/10.1145/3424636.3426894}
\showDOI{\tempurl}


\bibitem[\protect\citeauthoryear{He, Gao, Xiao, Zhang, Wang, Wang, Luo, He, Sobanbab, Pan, et~al\mbox{.}}{He et~al\mbox{.}}{2025a}]%
        {he2025asap}
\bibfield{author}{\bibinfo{person}{Tairan He}, \bibinfo{person}{Jiawei Gao}, \bibinfo{person}{Wenli Xiao}, \bibinfo{person}{Yuanhang Zhang}, \bibinfo{person}{Zi Wang}, \bibinfo{person}{Jiashun Wang}, \bibinfo{person}{Zhengyi Luo}, \bibinfo{person}{Guanqi He}, \bibinfo{person}{Nikhil Sobanbab}, \bibinfo{person}{Chaoyi Pan}, {et~al\mbox{.}}} \bibinfo{year}{2025}\natexlab{a}.
\newblock \showarticletitle{Asap: Aligning simulation and real-world physics for learning agile humanoid whole-body skills}.
\newblock \bibinfo{journal}{\emph{arXiv preprint arXiv:2502.01143}} (\bibinfo{year}{2025}).
\newblock


\bibitem[\protect\citeauthoryear{He, Luo, He, Xiao, Zhang, Zhang, Kitani, Liu, and Shi}{He et~al\mbox{.}}{2024}]%
        {he2024omnih2o}
\bibfield{author}{\bibinfo{person}{Tairan He}, \bibinfo{person}{Zhengyi Luo}, \bibinfo{person}{Xialin He}, \bibinfo{person}{Wenli Xiao}, \bibinfo{person}{Chong Zhang}, \bibinfo{person}{Weinan Zhang}, \bibinfo{person}{Kris Kitani}, \bibinfo{person}{Changliu Liu}, {and} \bibinfo{person}{Guanya Shi}.} \bibinfo{year}{2024}\natexlab{}.
\newblock \showarticletitle{Omnih2o: Universal and dexterous human-to-humanoid whole-body teleoperation and learning}.
\newblock \bibinfo{journal}{\emph{arXiv preprint arXiv:2406.08858}} (\bibinfo{year}{2024}).
\newblock


\bibitem[\protect\citeauthoryear{He, Xiao, Lin, Luo, Xu, Jiang, Kautz, Liu, Shi, Wang, et~al\mbox{.}}{He et~al\mbox{.}}{2025b}]%
        {he2025hover}
\bibfield{author}{\bibinfo{person}{Tairan He}, \bibinfo{person}{Wenli Xiao}, \bibinfo{person}{Toru Lin}, \bibinfo{person}{Zhengyi Luo}, \bibinfo{person}{Zhenjia Xu}, \bibinfo{person}{Zhenyu Jiang}, \bibinfo{person}{Jan Kautz}, \bibinfo{person}{Changliu Liu}, \bibinfo{person}{Guanya Shi}, \bibinfo{person}{Xiaolong Wang}, {et~al\mbox{.}}} \bibinfo{year}{2025}\natexlab{b}.
\newblock \showarticletitle{Hover: Versatile neural whole-body controller for humanoid robots}. In \bibinfo{booktitle}{\emph{2025 IEEE International Conference on Robotics and Automation (ICRA)}}. IEEE, \bibinfo{pages}{9989--9996}.
\newblock


\bibitem[\protect\citeauthoryear{Holden, Komura, and Saito}{Holden et~al\mbox{.}}{2017}]%
        {10.1145/3072959.3073663}
\bibfield{author}{\bibinfo{person}{Daniel Holden}, \bibinfo{person}{Taku Komura}, {and} \bibinfo{person}{Jun Saito}.} \bibinfo{year}{2017}\natexlab{}.
\newblock \showarticletitle{Phase-functioned neural networks for character control}.
\newblock \bibinfo{journal}{\emph{ACM Trans. Graph.}} \bibinfo{volume}{36}, \bibinfo{number}{4}, Article \bibinfo{articleno}{42} (\bibinfo{date}{July} \bibinfo{year}{2017}), \bibinfo{numpages}{13}~pages.
\newblock
\showISSN{0730-0301}
\urldef\tempurl%
\url{https://doi.org/10.1145/3072959.3073663}
\showDOI{\tempurl}


\bibitem[\protect\citeauthoryear{Ji, Peng, Liu, Li, Yang, Cheng, and Wang}{Ji et~al\mbox{.}}{2024}]%
        {ji2024exbody2}
\bibfield{author}{\bibinfo{person}{Mazeyu Ji}, \bibinfo{person}{Xuanbin Peng}, \bibinfo{person}{Fangchen Liu}, \bibinfo{person}{Jialong Li}, \bibinfo{person}{Ge Yang}, \bibinfo{person}{Xuxin Cheng}, {and} \bibinfo{person}{Xiaolong Wang}.} \bibinfo{year}{2024}\natexlab{}.
\newblock \showarticletitle{ExBody2: Advanced Expressive Humanoid Whole-Body Control}.
\newblock \bibinfo{journal}{\emph{arXiv preprint arXiv:2412.13196}} (\bibinfo{year}{2024}).
\newblock


\bibitem[\protect\citeauthoryear{Kulkarni, Rempe, Genova, Kundu, Johnson, Fouhey, and Guibas}{Kulkarni et~al\mbox{.}}{2024}]%
        {kulkarni2024nifty}
\bibfield{author}{\bibinfo{person}{Nilesh Kulkarni}, \bibinfo{person}{Davis Rempe}, \bibinfo{person}{Kyle Genova}, \bibinfo{person}{Abhijit Kundu}, \bibinfo{person}{Justin Johnson}, \bibinfo{person}{David Fouhey}, {and} \bibinfo{person}{Leonidas Guibas}.} \bibinfo{year}{2024}\natexlab{}.
\newblock \showarticletitle{Nifty: Neural object interaction fields for guided human motion synthesis}. In \bibinfo{booktitle}{\emph{Proceedings of the IEEE/CVF Conference on Computer Vision and Pattern Recognition}}. \bibinfo{pages}{947--957}.
\newblock


\bibitem[\protect\citeauthoryear{Kundu, Buckchash, Mandikal, V, Jamkhandi, and Babu}{Kundu et~al\mbox{.}}{2020}]%
        {9093627}
\bibfield{author}{\bibinfo{person}{Jogendra~Nath Kundu}, \bibinfo{person}{Himanshu Buckchash}, \bibinfo{person}{Priyanka Mandikal}, \bibinfo{person}{Rahul~M V}, \bibinfo{person}{Anirudh Jamkhandi}, {and} \bibinfo{person}{R.~Venkatesh Babu}.} \bibinfo{year}{2020}\natexlab{}.
\newblock \showarticletitle{Cross-Conditioned Recurrent Networks for Long-Term Synthesis of Inter-Person Human Motion Interactions}. In \bibinfo{booktitle}{\emph{2020 IEEE Winter Conference on Applications of Computer Vision (WACV)}}. \bibinfo{pages}{2713--2722}.
\newblock
\urldef\tempurl%
\url{https://doi.org/10.1109/WACV45572.2020.9093627}
\showDOI{\tempurl}


\bibitem[\protect\citeauthoryear{Li, Chibane, He, Pearl, Geiger, and Pons-Moll}{Li et~al\mbox{.}}{2025}]%
        {li2025unimotion}
\bibfield{author}{\bibinfo{person}{Chuqiao Li}, \bibinfo{person}{Julian Chibane}, \bibinfo{person}{Yannan He}, \bibinfo{person}{Naama Pearl}, \bibinfo{person}{Andreas Geiger}, {and} \bibinfo{person}{Gerard Pons-Moll}.} \bibinfo{year}{2025}\natexlab{}.
\newblock \showarticletitle{Unimotion: Unifying 3d human motion synthesis and understanding}. In \bibinfo{booktitle}{\emph{2025 International Conference on 3D Vision (3DV)}}. IEEE, \bibinfo{pages}{240--249}.
\newblock


\bibitem[\protect\citeauthoryear{Liang, Zhang, Li, Yu, and Xu}{Liang et~al\mbox{.}}{2024}]%
        {Liang_2024}
\bibfield{author}{\bibinfo{person}{Han Liang}, \bibinfo{person}{Wenqian Zhang}, \bibinfo{person}{Wenxuan Li}, \bibinfo{person}{Jingyi Yu}, {and} \bibinfo{person}{Lan Xu}.} \bibinfo{year}{2024}\natexlab{}.
\newblock \showarticletitle{InterGen: Diffusion-Based Multi-human Motion Generation Under Complex Interactions}.
\newblock \bibinfo{journal}{\emph{International Journal of Computer Vision}} \bibinfo{volume}{132}, \bibinfo{number}{9} (\bibinfo{date}{March} \bibinfo{year}{2024}), \bibinfo{pages}{3463–3483}.
\newblock
\showISSN{1573-1405}
\urldef\tempurl%
\url{https://doi.org/10.1007/s11263-024-02042-6}
\showDOI{\tempurl}


\bibitem[\protect\citeauthoryear{Liao, Truong, Huang, Tevet, Sreenath, and Liu}{Liao et~al\mbox{.}}{2025}]%
        {liao2025beyondmimicmotiontrackingversatile}
\bibfield{author}{\bibinfo{person}{Qiayuan Liao}, \bibinfo{person}{Takara~E. Truong}, \bibinfo{person}{Xiaoyu Huang}, \bibinfo{person}{Guy Tevet}, \bibinfo{person}{Koushil Sreenath}, {and} \bibinfo{person}{C.~Karen Liu}.} \bibinfo{year}{2025}\natexlab{}.
\newblock \bibinfo{title}{BeyondMimic: From Motion Tracking to Versatile Humanoid Control via Guided Diffusion}.
\newblock
\newblock
\showeprint[arxiv]{2508.08241}~[cs.RO]
\urldef\tempurl%
\url{https://arxiv.org/abs/2508.08241}
\showURL{%
\tempurl}


\bibitem[\protect\citeauthoryear{Luo, Cao, Winkler, Kitani, and Xu}{Luo et~al\mbox{.}}{2023}]%
        {luo2023perpetualhumanoidcontrolrealtime}
\bibfield{author}{\bibinfo{person}{Zhengyi Luo}, \bibinfo{person}{Jinkun Cao}, \bibinfo{person}{Alexander Winkler}, \bibinfo{person}{Kris Kitani}, {and} \bibinfo{person}{Weipeng Xu}.} \bibinfo{year}{2023}\natexlab{}.
\newblock \bibinfo{title}{Perpetual Humanoid Control for Real-time Simulated Avatars}.
\newblock
\newblock
\showeprint[arxiv]{2305.06456}~[cs.CV]
\urldef\tempurl%
\url{https://arxiv.org/abs/2305.06456}
\showURL{%
\tempurl}


\bibitem[\protect\citeauthoryear{Mahmood, Ghorbani, Troje, Pons-Moll, and Black}{Mahmood et~al\mbox{.}}{2019}]%
        {AMASS:ICCV:2019}
\bibfield{author}{\bibinfo{person}{Naureen Mahmood}, \bibinfo{person}{Nima Ghorbani}, \bibinfo{person}{Nikolaus~F. Troje}, \bibinfo{person}{Gerard Pons-Moll}, {and} \bibinfo{person}{Michael~J. Black}.} \bibinfo{year}{2019}\natexlab{}.
\newblock \showarticletitle{{AMASS}: Archive of Motion Capture as Surface Shapes}. In \bibinfo{booktitle}{\emph{International Conference on Computer Vision}}. \bibinfo{pages}{5442--5451}.
\newblock


\bibitem[\protect\citeauthoryear{Makoviychuk, Wawrzyniak, Guo, Lu, Storey, Macklin, Hoeller, Rudin, Allshire, Handa, et~al\mbox{.}}{Makoviychuk et~al\mbox{.}}{2021}]%
        {makoviychuk2021isaac}
\bibfield{author}{\bibinfo{person}{Viktor Makoviychuk}, \bibinfo{person}{Lukasz Wawrzyniak}, \bibinfo{person}{Yunrong Guo}, \bibinfo{person}{Michelle Lu}, \bibinfo{person}{Kier Storey}, \bibinfo{person}{Miles Macklin}, \bibinfo{person}{David Hoeller}, \bibinfo{person}{Nikita Rudin}, \bibinfo{person}{Arthur Allshire}, \bibinfo{person}{Ankur Handa}, {et~al\mbox{.}}} \bibinfo{year}{2021}\natexlab{}.
\newblock \showarticletitle{Isaac gym: High performance gpu-based physics simulation for robot learning}.
\newblock \bibinfo{journal}{\emph{arXiv preprint arXiv:2108.10470}} (\bibinfo{year}{2021}).
\newblock


\bibitem[\protect\citeauthoryear{Mittal, Yu, Yu, Liu, Rudin, Hoeller, Yuan, Singh, Guo, Mazhar, Mandlekar, Babich, State, Hutter, and Garg}{Mittal et~al\mbox{.}}{2023}]%
        {mittal2023orbit}
\bibfield{author}{\bibinfo{person}{Mayank Mittal}, \bibinfo{person}{Calvin Yu}, \bibinfo{person}{Qinxi Yu}, \bibinfo{person}{Jingzhou Liu}, \bibinfo{person}{Nikita Rudin}, \bibinfo{person}{David Hoeller}, \bibinfo{person}{Jia~Lin Yuan}, \bibinfo{person}{Ritvik Singh}, \bibinfo{person}{Yunrong Guo}, \bibinfo{person}{Hammad Mazhar}, \bibinfo{person}{Ajay Mandlekar}, \bibinfo{person}{Buck Babich}, \bibinfo{person}{Gavriel State}, \bibinfo{person}{Marco Hutter}, {and} \bibinfo{person}{Animesh Garg}.} \bibinfo{year}{2023}\natexlab{}.
\newblock \showarticletitle{Orbit: A Unified Simulation Framework for Interactive Robot Learning Environments}.
\newblock \bibinfo{journal}{\emph{IEEE Robotics and Automation Letters}} \bibinfo{volume}{8}, \bibinfo{number}{6} (\bibinfo{year}{2023}), \bibinfo{pages}{3740--3747}.
\newblock
\urldef\tempurl%
\url{https://doi.org/10.1109/LRA.2023.3270034}
\showDOI{\tempurl}


\bibitem[\protect\citeauthoryear{Peng, Abbeel, Levine, and van~de Panne}{Peng et~al\mbox{.}}{2018}]%
        {2018-TOG-deepMimic}
\bibfield{author}{\bibinfo{person}{Xue~Bin Peng}, \bibinfo{person}{Pieter Abbeel}, \bibinfo{person}{Sergey Levine}, {and} \bibinfo{person}{Michiel van~de Panne}.} \bibinfo{year}{2018}\natexlab{}.
\newblock \showarticletitle{DeepMimic: Example-guided Deep Reinforcement Learning of Physics-based Character Skills}.
\newblock \bibinfo{journal}{\emph{ACM Trans. Graph.}} \bibinfo{volume}{37}, \bibinfo{number}{4}, Article \bibinfo{articleno}{143} (\bibinfo{date}{July} \bibinfo{year}{2018}), \bibinfo{numpages}{14}~pages.
\newblock
\showISSN{0730-0301}
\urldef\tempurl%
\url{https://doi.org/10.1145/3197517.3201311}
\showDOI{\tempurl}


\bibitem[\protect\citeauthoryear{Rudin, Hoeller, Reist, and Hutter}{Rudin et~al\mbox{.}}{2022}]%
        {rudin2022learningwalkminutesusing}
\bibfield{author}{\bibinfo{person}{Nikita Rudin}, \bibinfo{person}{David Hoeller}, \bibinfo{person}{Philipp Reist}, {and} \bibinfo{person}{Marco Hutter}.} \bibinfo{year}{2022}\natexlab{}.
\newblock \bibinfo{title}{Learning to Walk in Minutes Using Massively Parallel Deep Reinforcement Learning}.
\newblock
\newblock
\showeprint[arxiv]{2109.11978}~[cs.RO]
\urldef\tempurl%
\url{https://arxiv.org/abs/2109.11978}
\showURL{%
\tempurl}


\bibitem[\protect\citeauthoryear{Schulman, Wolski, Dhariwal, Radford, and Klimov}{Schulman et~al\mbox{.}}{2017}]%
        {schulman2017proximal}
\bibfield{author}{\bibinfo{person}{John Schulman}, \bibinfo{person}{Filip Wolski}, \bibinfo{person}{Prafulla Dhariwal}, \bibinfo{person}{Alec Radford}, {and} \bibinfo{person}{Oleg Klimov}.} \bibinfo{year}{2017}\natexlab{}.
\newblock \showarticletitle{Proximal policy optimization algorithms}.
\newblock \bibinfo{journal}{\emph{arXiv preprint arXiv:1707.06347}} (\bibinfo{year}{2017}).
\newblock


\bibitem[\protect\citeauthoryear{Shafir, Tevet, Kapon, and Bermano}{Shafir et~al\mbox{.}}{2023}]%
        {shafir2023human}
\bibfield{author}{\bibinfo{person}{Yonatan Shafir}, \bibinfo{person}{Guy Tevet}, \bibinfo{person}{Roy Kapon}, {and} \bibinfo{person}{Amit~H Bermano}.} \bibinfo{year}{2023}\natexlab{}.
\newblock \showarticletitle{Human motion diffusion as a generative prior}.
\newblock \bibinfo{journal}{\emph{arXiv preprint arXiv:2303.01418}} (\bibinfo{year}{2023}).
\newblock


\bibitem[\protect\citeauthoryear{Starke, Zhang, Komura, and Saito}{Starke et~al\mbox{.}}{2019}]%
        {10.1145/3355089.3356505}
\bibfield{author}{\bibinfo{person}{Sebastian Starke}, \bibinfo{person}{He Zhang}, \bibinfo{person}{Taku Komura}, {and} \bibinfo{person}{Jun Saito}.} \bibinfo{year}{2019}\natexlab{}.
\newblock \showarticletitle{Neural state machine for character-scene interactions}.
\newblock \bibinfo{journal}{\emph{ACM Trans. Graph.}} \bibinfo{volume}{38}, \bibinfo{number}{6}, Article \bibinfo{articleno}{209} (\bibinfo{date}{Nov.} \bibinfo{year}{2019}), \bibinfo{numpages}{14}~pages.
\newblock
\showISSN{0730-0301}
\urldef\tempurl%
\url{https://doi.org/10.1145/3355089.3356505}
\showDOI{\tempurl}


\bibitem[\protect\citeauthoryear{Starke, Zhao, Komura, and Zaman}{Starke et~al\mbox{.}}{2020}]%
        {10.1145/3386569.3392450}
\bibfield{author}{\bibinfo{person}{Sebastian Starke}, \bibinfo{person}{Yiwei Zhao}, \bibinfo{person}{Taku Komura}, {and} \bibinfo{person}{Kazi Zaman}.} \bibinfo{year}{2020}\natexlab{}.
\newblock \showarticletitle{Local motion phases for learning multi-contact character movements}.
\newblock \bibinfo{journal}{\emph{ACM Trans. Graph.}} \bibinfo{volume}{39}, \bibinfo{number}{4}, Article \bibinfo{articleno}{54} (\bibinfo{date}{Aug.} \bibinfo{year}{2020}), \bibinfo{numpages}{14}~pages.
\newblock
\showISSN{0730-0301}
\urldef\tempurl%
\url{https://doi.org/10.1145/3386569.3392450}
\showDOI{\tempurl}


\bibitem[\protect\citeauthoryear{Tobin, Fong, Ray, Schneider, Zaremba, and Abbeel}{Tobin et~al\mbox{.}}{2017}]%
        {tobin2017domain}
\bibfield{author}{\bibinfo{person}{Josh Tobin}, \bibinfo{person}{Rachel Fong}, \bibinfo{person}{Alex Ray}, \bibinfo{person}{Jonas Schneider}, \bibinfo{person}{Wojciech Zaremba}, {and} \bibinfo{person}{Pieter Abbeel}.} \bibinfo{year}{2017}\natexlab{}.
\newblock \showarticletitle{Domain randomization for transferring deep neural networks from simulation to the real world}. In \bibinfo{booktitle}{\emph{2017 IEEE/RSJ international conference on intelligent robots and systems (IROS)}}. IEEE, \bibinfo{pages}{23--30}.
\newblock


\bibitem[\protect\citeauthoryear{Weng, Li, Sobanbabu, Wang, Luo, He, Ramanan, and Shi}{Weng et~al\mbox{.}}{2025}]%
        {weng2025hdmilearninginteractivehumanoid}
\bibfield{author}{\bibinfo{person}{Haoyang Weng}, \bibinfo{person}{Yitang Li}, \bibinfo{person}{Nikhil Sobanbabu}, \bibinfo{person}{Zihan Wang}, \bibinfo{person}{Zhengyi Luo}, \bibinfo{person}{Tairan He}, \bibinfo{person}{Deva Ramanan}, {and} \bibinfo{person}{Guanya Shi}.} \bibinfo{year}{2025}\natexlab{}.
\newblock \bibinfo{title}{HDMI: Learning Interactive Humanoid Whole-Body Control from Human Videos}.
\newblock
\newblock
\showeprint[arxiv]{2509.16757}~[cs.RO]
\urldef\tempurl%
\url{https://arxiv.org/abs/2509.16757}
\showURL{%
\tempurl}


\bibitem[\protect\citeauthoryear{Xie, Han, Zheng, Li, Liu, Shi, Zhang, Bai, and Li}{Xie et~al\mbox{.}}{2025}]%
        {xie2025kungfubot}
\bibfield{author}{\bibinfo{person}{Weiji Xie}, \bibinfo{person}{Jinrui Han}, \bibinfo{person}{Jiakun Zheng}, \bibinfo{person}{Huanyu Li}, \bibinfo{person}{Xinzhe Liu}, \bibinfo{person}{Jiyuan Shi}, \bibinfo{person}{Weinan Zhang}, \bibinfo{person}{Chenjia Bai}, {and} \bibinfo{person}{Xuelong Li}.} \bibinfo{year}{2025}\natexlab{}.
\newblock \showarticletitle{KungfuBot: Physics-Based Humanoid Whole-Body Control for Learning Highly-Dynamic Skills}.
\newblock \bibinfo{journal}{\emph{arXiv preprint arXiv:2506.12851}} (\bibinfo{year}{2025}).
\newblock


\bibitem[\protect\citeauthoryear{Xu, Lv, Yan, Jin, Wu, Xu, Liu, Zhou, Rao, Sheng, et~al\mbox{.}}{Xu et~al\mbox{.}}{2024a}]%
        {xu2024inter}
\bibfield{author}{\bibinfo{person}{Liang Xu}, \bibinfo{person}{Xintao Lv}, \bibinfo{person}{Yichao Yan}, \bibinfo{person}{Xin Jin}, \bibinfo{person}{Shuwen Wu}, \bibinfo{person}{Congsheng Xu}, \bibinfo{person}{Yifan Liu}, \bibinfo{person}{Yizhou Zhou}, \bibinfo{person}{Fengyun Rao}, \bibinfo{person}{Xingdong Sheng}, {et~al\mbox{.}}} \bibinfo{year}{2024}\natexlab{a}.
\newblock \showarticletitle{Inter-x: Towards versatile human-human interaction analysis}. In \bibinfo{booktitle}{\emph{Proceedings of the IEEE/CVF conference on computer vision and pattern recognition}}. \bibinfo{pages}{22260--22271}.
\newblock


\bibitem[\protect\citeauthoryear{Xu, Zhou, Yan, Jin, Zhu, Rao, Yang, and Zeng}{Xu et~al\mbox{.}}{2024b}]%
        {xu2024regennethumanactionreactionsynthesis}
\bibfield{author}{\bibinfo{person}{Liang Xu}, \bibinfo{person}{Yizhou Zhou}, \bibinfo{person}{Yichao Yan}, \bibinfo{person}{Xin Jin}, \bibinfo{person}{Wenhan Zhu}, \bibinfo{person}{Fengyun Rao}, \bibinfo{person}{Xiaokang Yang}, {and} \bibinfo{person}{Wenjun Zeng}.} \bibinfo{year}{2024}\natexlab{b}.
\newblock \bibinfo{title}{ReGenNet: Towards Human Action-Reaction Synthesis}.
\newblock
\newblock
\showeprint[arxiv]{2403.11882}~[cs.CV]
\urldef\tempurl%
\url{https://arxiv.org/abs/2403.11882}
\showURL{%
\tempurl}


\bibitem[\protect\citeauthoryear{Yang, Huang, Wu, Kanazawa, Abbeel, Sferrazza, Liu, Duan, and Shi}{Yang et~al\mbox{.}}{2025}]%
        {yang2025omniretargetinteractionpreservingdatageneration}
\bibfield{author}{\bibinfo{person}{Lujie Yang}, \bibinfo{person}{Xiaoyu Huang}, \bibinfo{person}{Zhen Wu}, \bibinfo{person}{Angjoo Kanazawa}, \bibinfo{person}{Pieter Abbeel}, \bibinfo{person}{Carmelo Sferrazza}, \bibinfo{person}{C.~Karen Liu}, \bibinfo{person}{Rocky Duan}, {and} \bibinfo{person}{Guanya Shi}.} \bibinfo{year}{2025}\natexlab{}.
\newblock \bibinfo{title}{OmniRetarget: Interaction-Preserving Data Generation for Humanoid Whole-Body Loco-Manipulation and Scene Interaction}.
\newblock
\newblock
\showeprint[arxiv]{2509.26633}~[cs.RO]
\urldef\tempurl%
\url{https://arxiv.org/abs/2509.26633}
\showURL{%
\tempurl}


\bibitem[\protect\citeauthoryear{Yin, Zeng, Fan, Wang, Zhang, Tian, Wang, Pang, and Zhang}{Yin et~al\mbox{.}}{2025b}]%
        {yin2025unitracker}
\bibfield{author}{\bibinfo{person}{Kangning Yin}, \bibinfo{person}{Weishuai Zeng}, \bibinfo{person}{Ke Fan}, \bibinfo{person}{Zirui Wang}, \bibinfo{person}{Qiang Zhang}, \bibinfo{person}{Zheng Tian}, \bibinfo{person}{Jingbo Wang}, \bibinfo{person}{Jiangmiao Pang}, {and} \bibinfo{person}{Weinan Zhang}.} \bibinfo{year}{2025}\natexlab{b}.
\newblock \showarticletitle{Unitracker: Learning universal whole-body motion tracker for humanoid robots}.
\newblock \bibinfo{journal}{\emph{arXiv preprint arXiv:2507.07356}} (\bibinfo{year}{2025}).
\newblock


\bibitem[\protect\citeauthoryear{Yin, Ze, Yu, Liu, and Wu}{Yin et~al\mbox{.}}{2025a}]%
        {yin2025visualmimicvisualhumanoidlocomanipulation}
\bibfield{author}{\bibinfo{person}{Shaofeng Yin}, \bibinfo{person}{Yanjie Ze}, \bibinfo{person}{Hong-Xing Yu}, \bibinfo{person}{C.~Karen Liu}, {and} \bibinfo{person}{Jiajun Wu}.} \bibinfo{year}{2025}\natexlab{a}.
\newblock \bibinfo{title}{VisualMimic: Visual Humanoid Loco-Manipulation via Motion Tracking and Generation}.
\newblock
\newblock
\showeprint[arxiv]{2509.20322}~[cs.RO]
\urldef\tempurl%
\url{https://arxiv.org/abs/2509.20322}
\showURL{%
\tempurl}


\bibitem[\protect\citeauthoryear{Zakka}{Zakka}{2025}]%
        {Zakka_Mink_Python_inverse_2025}
\bibfield{author}{\bibinfo{person}{Kevin Zakka}.} \bibinfo{year}{2025}\natexlab{}.
\newblock \bibinfo{booktitle}{\emph{{Mink: Python inverse kinematics based on MuJoCo}}}.
\newblock
\urldef\tempurl%
\url{https://github.com/kevinzakka/mink}
\showURL{%
\tempurl}


\bibitem[\protect\citeauthoryear{Ze, Araújo, Wu, and Liu}{Ze et~al\mbox{.}}{2025a}]%
        {ze2025gmr}
\bibfield{author}{\bibinfo{person}{Yanjie Ze}, \bibinfo{person}{João~Pedro Araújo}, \bibinfo{person}{Jiajun Wu}, {and} \bibinfo{person}{C.~Karen Liu}.} \bibinfo{year}{2025}\natexlab{a}.
\newblock \bibinfo{booktitle}{\emph{GMR: General Motion Retargeting}}.
\newblock
\urldef\tempurl%
\url{https://github.com/YanjieZe/GMR}
\showURL{%
\tempurl}
\newblock
\shownote{GitHub repository.}


\bibitem[\protect\citeauthoryear{Ze, Chen, Araújo, ang Cao, Peng, Wu, and Liu}{Ze et~al\mbox{.}}{2025b}]%
        {ze2025twist}
\bibfield{author}{\bibinfo{person}{Yanjie Ze}, \bibinfo{person}{Zixuan Chen}, \bibinfo{person}{João~Pedro Araújo}, \bibinfo{person}{Zi ang Cao}, \bibinfo{person}{Xue~Bin Peng}, \bibinfo{person}{Jiajun Wu}, {and} \bibinfo{person}{C.~Karen Liu}.} \bibinfo{year}{2025}\natexlab{b}.
\newblock \showarticletitle{TWIST: Teleoperated Whole-Body Imitation System}.
\newblock \bibinfo{journal}{\emph{arXiv preprint arXiv:2505.02833}} (\bibinfo{year}{2025}).
\newblock


\bibitem[\protect\citeauthoryear{Zhang, Gopinath, Ye, Hodgins, Turk, and Won}{Zhang et~al\mbox{.}}{2023}]%
        {zhang2023simulationretargetingcomplexmulticharacter}
\bibfield{author}{\bibinfo{person}{Yunbo Zhang}, \bibinfo{person}{Deepak Gopinath}, \bibinfo{person}{Yuting Ye}, \bibinfo{person}{Jessica Hodgins}, \bibinfo{person}{Greg Turk}, {and} \bibinfo{person}{Jungdam Won}.} \bibinfo{year}{2023}\natexlab{}.
\newblock \showarticletitle{Simulation and retargeting of complex multi-character interactions}. In \bibinfo{booktitle}{\emph{ACM SIGGRAPH 2023 Conference Proceedings}}. \bibinfo{pages}{1--11}.
\newblock


\end{thebibliography}
